\documentclass[ijoc,nonblindrev]{informs3} 

\DoubleSpacedXII

\usepackage{threeparttable}
\usepackage{longtable}
\usepackage{booktabs}
\usepackage{url}
\usepackage[colorlinks=true,linkcolor=blue,citecolor=blue,urlcolor=blue]{hyperref}
\usepackage{caption}
\usepackage{multirow}
\usepackage[ruled,linesnumbered]{algorithm2e}

\usepackage{subfigure}
\usepackage[table]{xcolor}
\usepackage{amssymb}
\usepackage{graphicx}
\usepackage{float}
\usepackage{lineno}

\usepackage{natbib}
 \bibpunct[, ]{(}{)}{,}{a}{}{,}%
 \def\bibfont{\small}%
\TheoremsNumberedThrough     

\EquationsNumberedThrough    

\def\TheoremsNumberedThrough{%
\theoremstyle{TH}%
\newtheorem{theorem}{Theorem}
\newtheorem{lemma}{Lemma}
\newtheorem{proposition}{Proposition}
\newtheorem{corollary}{Corollary}
\newtheorem{claim}{Claim}
\newtheorem{conjecture}{Conjecture}
\newtheorem{hypothesis}{Hypothesis}
\newtheorem{assumption}{Assumption}
\theoremstyle{EX}
\newtheorem{remark}{Remark}
\newtheorem{example}{Example}
\newtheorem{problem}{Problem}
\newtheorem{definition}{Definition}
\newtheorem{question}{Question}
\newtheorem{answer}{Answer}
\newtheorem{exercise}{Exercise}
}

\usepackage[utf8]{inputenc}

\usepackage{amssymb}
\usepackage{float}
\usepackage{lineno}
\usepackage{comment}
\usepackage{natbib}
 \bibpunct[, ]{(}{)}{,}{a}{}{,}%
 \def\bibfont{\small}%
\EquationsNumberedThrough    

\def\TheoremsNumberedThrough{%
\theoremstyle{TH}%

\theoremstyle{EX}

}
\begin{document}


 \RUNAUTHOR{He, Hao, and Wu}

\RUNTITLE{A Multi-population Integrated Approach for CLRP}

\TITLE{A Multi-population Integrated Approach for Capacitated Location Routing}

\ARTICLEAUTHORS{%
\AUTHOR{Pengfei He}
\AFF{School of Automation and Key Laboratory of Measurement and Control of Complex Systems of Engineering, Ministry of Education, Southeast University, Nanjing 210096, China.\\ \EMAIL{pengfeihe606@gmail.com}}
\AUTHOR{Jin-Kao Hao*}
\AFF{Department of Computer Science, LERIA, University of Angers, 2 Boulevard Lavoisier, 49045 Angers, France\\ \EMAIL{jin-kao.hao@univ-angers.fr}}
\AUTHOR{Qinghua Wu}
\AFF{School of Management, Huazhong University of Science and Technology, No. 1037, Luoyu Road, Wuhan, China.\\
\EMAIL{qinghuawu1005@gmail.com}}

} 

\ABSTRACT{%
The capacitated location-routing problem involves determining the depots from a set of candidate  capacitated depot locations and finding the required routes from the selected depots to serve a set of customers whereas minimizing a cost function that includes the cost of opening the chosen depots, the fixed utilization cost per vehicle used, and the total cost (distance) of the routes. This paper presents a multi-population integrated framework in which a multi-depot edge assembly crossover generates promising offspring solutions from the perspective of both depot location and route edge assembly. The method includes an effective neighborhood-based local search, a feasibility-restoring procedure and a diversification-oriented mutation. Of particular interest is the multi-population scheme which organizes the population into multiple subpopulations based on depot configurations. Extensive experiments on 281 benchmark instances from the literature show that the algorithm performs remarkably well, by improving 101 best-known results (new upper bounds) and matching 84 best-known results. Additional experiments are presented to gain insight into the role of the key elements of the algorithm.
}


\maketitle

%


\section{Introduction}\label{intro}

Let $\mathcal{I} = \{1,\cdots, m\}$ be a set of depot locations (or simply depots), where each depot $i \in \mathcal{I}$ is associated with a positive capacity $w_i$ and an opening cost $o_i$. Each depot has an unlimited fleet of vehicles. Each vehicle has a limited capacity $Q$ and a fixed utilization cost $F$. Let $\mathcal{J}= \{1,\cdots,n\}$ be a set of customers, where each customer has a positive demand $d_j$. The capacitated location-routing problem (CLRP) is defined on a weighted and directed graph $\mathcal{G=(V,A)}$ with vertex set $\mathcal{V} = \mathcal{I} \cup \mathcal{J}$ and arc set $\mathcal{A} = \{(i,j)|i\in\mathcal{I}, j\in \mathcal{J}\} \cup \{(i,j)|i,j\in \mathcal{J}, i \neq j\} \cup \{(i,j)|i\in\mathcal{J}, j\in \mathcal{I}\}$. $\mathcal{A}$ is associated with a matrix $\mathcal{C}=(c_{ij})$ where $c_{ij}$ is a non-negative value representing the distance of the arc $(i,j)$. The distance matrix $\mathcal{C}$ is said to be symmetric when $c_{ij} = c_{ji}$, for each $(i,j)\in \mathcal{A}$ and asymmetric otherwise. Let $\mathcal{E}$ be an edge set and $\mathcal{E} = \mathcal{A}$ if $\mathcal{G}$ is undirected. The CLRP problem involves determining the depots to open and the routes from these depots to visit all customers under the following constraints: (i) vehicle and depot capacities are respected; (ii) each vehicle ends at the depot from which it originated; (iii) each customer is visited exactly once. The objective of the CLRP is to minimize the total cost, including the cost of opening the chosen depots, the fixed utilization cost per vehicle used, and the total cost (distance) of the routes. 

The CLRP covers a variety of problems that arise in single and two-echelon distribution networks in urban logistics \citep{prodhon2014survey}. On the one hand, once the opening depots are fixed, the CLRP reduces to the $\mathcal{NP}$-hard multi-depot vehicle routing problem (MDVRP) \citep{cordeau1997tabu}. On the other hand, locations in the CLRP correspond to satellites in the two-echelon vehicle routing problem (2E-VRP) \citep{perboli2011two}. The relationship between these three problems has been analyzed by \citet{voigt2022hybrid} and \citet{schneider2019large}. Moreover, the CLRP represents a well-studied class of vehicle routing problems (VRPs) in which decisions about depots and routes are associated. Many variations of the CLRP arise in freight distribution and urban logistics with the addition of specific constraints that model practical scenarios \citep{drexl2015survey,mara2021location}. For a mathematical formulation of the CLRP, please refer \citet{baldacci2011exact}. 

Given the computational challenge and practical relevance of the CLRP \citep{nagy2007location}, much effort has been devoted to the development of efficient solution methods to better solve the problem. The three most representative exact algorithms \citep{contardo2014exact,baldacci2011exact,liguori2022non}, based on the branch-and-cut-and-price framework, are able to optimally solve instances with up to 200 customers and 10 depots at the cost of high computational times (e.g., up to 176 hours for instances with up to 150/199 customers and 14 depots \citep{contardo2014exact}). To handle larger instances, a number of heuristic algorithms have been proposed that aim to find high-quality solutions in a reasonable amount of time. In Section \ref{liter_rev}, we give an overview  of the main existing heuristics, which follow two general approaches: the hierarchical approach and the integrated approach. The hierarchical approach first determines a depot configuration and then uses a routing procedure to solve the MDVRP resulting from the depot configuration, whereas the integrated approach simultaneously handles the location decision and the route optimization together. As discussed in Section \ref{liter_rev}, whereas these approaches have their merits, both of them face significant challenges. In particular, for the hierarchical approach, it is a difficult task to identify good depot configurations, which are critical to the subsequent routing task and strongly influence the final solution. For the integrated approach, the simultaneous optimization of the depot configuration and the routing implies the exploration of a larger search space and presents a real challenge to the search procedure. 

In this work, we propose a hybrid genetic algorithm with multi-population (HGAMP) that combines a multi-depot edge assembly crossover (mdEAX) to dynamically explore promising depot configurations and an effective neighborhood-based local optimization to perform route optimization. In particular, our approach maintains an ensemble of elite solutions (routes) organized into multiple subpopulations, where each subpopulation consists of a set of high-quality solutions that share the same depot configuration. The mdEAX crossover has the ability to generate promising offspring solutions with new promising depot configurations, which is achieved by recombining route edges associated with different depot configurations. The method also uses a coverage ratio heuristic (CRH) to generate initial depot configurations by considering the best balance between cost efficiency and geographic distribution. Finally, the method takes full advantage of existing routing heuristics for effective routing optimization with a given depot configuration and a dedicated mutation to maintain high solution diversity.

We evaluate the performance of our algorithm on a total of 281 benchmark instances, including the three popular sets of 79 classical instances introduced in  \citet{tuzun1999two,prins2006solving,barreto2007using} and the recent set of 202 new instances with up to 600 customers and 30 depots introduced in   \citet{schneider2019large}. For the 79 classical instances, our algorithm achieved three new best results, despite the fact that these instances have been extensively used to evaluate various algorithms and their results are considered challenging to improve upon. For the 202 new instances, our algorithm significantly outperformed the existing methods by reporting improved best results for 98 instances. To gain insight into the performance and behavior of the algorithm, we perform additional experiments to verify the benefits of two essential algorithmic components, i.e., the multi-population scheme and the multi-depot edge assembly crossover.



In Section \ref{liter_rev}, we provide a literature review on the CLRP. In Section \ref{memetic_algo}, we present the proposed algorithm. In Section \ref{com_results}, we experimentally compare our algorithm with the state-of-the-art methods in the literature. In Section \ref{analysis}, we conduct additional experiments to analyze the main component of the algorithm. Finally, Section \ref{conclusion} concludes the paper.

\section{Literature review}\label{liter_rev}

This section reviews the representative heuristic algorithms for the CLRP. For a detailed presentation of the existing algorithms for the CLRP, the reader is referred to the surveys \citep{prodhon2014survey,schneider2017survey}. We organize the overview according to two solution approaches: the hierarchical approach and the integrated approach. The \textit{hierarchical approach} treats the CLRP in two sequential steps, where the first step identifies promising depot configurations and the second step solves the resulting MDVRP problems. The \textit{integrated approach} treats the location decision and the route optimization together. We summarize the main CLRP heuristics according to this classification in Table \ref{liter_table} and present the most representative algorithms in the rest of this section.

\begin{table}[]\caption{Representative heuristic algorithms for the CLRP.}\label{liter_table}
\begin{tiny}
\begin{center}
\begin{tabular}{p{3.5cm}p{0.6cm}p{3.4cm}p{1.4cm}p{0.8cm}p{1.2cm}p{0.01cm}p{0.7cm}p{0.7cm}p{1.0cm}}
\hline
&&&&\multicolumn{2}{c}{Strategy}&&\multicolumn{2}{c}{Capacity}\\
\cline{5-6}\cline{8-9}
\multirow{-2}{*}{Approaches} &\multirow{-2}{*}{Year}&\multirow{-2}{*}{Matheuristics/Metaheuristics}& \multirow{-2}{*}{Structure} & Divided &Simultaneous & & Routes        & Depots    &\multirow{-2}{*}{Instances} \\
\hline
\citet{tuzun1999two}   &1999      & Two phase tabu search&  Hierarchical       &       \checkmark       &               &&\checkmark               &   $\times$           & $\mathbb{T}$      \\
\citet{prins2006solving}      &2006    & GRASP + path relinking  & Hierarchical  & \checkmark             &               &&\checkmark               &      \checkmark        & $\mathbb{T,P,B}$          \\
\citet{prins2006memetic}       &2006                         & Memetic algorithm   &Integrated                             &              & \checkmark &               & \checkmark              &\checkmark              &       $\mathbb{T,P,B}$    \\
\citet{prins2007solving}       &2007                         & Lagrangean reaxation and TS & Hierarchical &   \checkmark                   && &\checkmark               &             \checkmark &         $\mathbb{T,P,B}$  \\
\citet{duhamel2010grasp}          &2010                      & GRASP+Evolutionary LS       &Integrated                    &     &\checkmark  &               &    \checkmark           &   \checkmark          &   $\mathbb{T,P,B}$        \\
\citet{vincent2010simulated}           &2010                 & Simulated annealing     &Integrated                        &      \checkmark        &               &&          \checkmark     &             \checkmark &     $\mathbb{T,P,B}$      \\
\citet{hemmelmayr2012adaptive}   &2012                       & ALNS         &Integrated                                   &    \checkmark          &&               &       \checkmark        &       \checkmark       &  $\mathbb{T,P,B}$         \\
\citet{ting2013multiple} & 2013 & Ant colony optimization	&	Hierarchical	& \checkmark			&&& \checkmark &					\checkmark & 	 $\mathbb{T,P,B}$																\\
\citet{escobar2013two}            &2013                      & Two phase hybrid heuristic &Hierarchical                       &    \checkmark          &               &&     \checkmark          &              \checkmark &        $\mathbb{T,P,B}$   \\
\citet{escobar2014granular}                &2014             & Granular tabu search         &Integrated                   &    \checkmark          &               &&       \checkmark        &    \checkmark          &       $\mathbb{T,P,B}$    \\
\citet{contardo2014grasp} &2014& GRASP + ILP     & Integrated & \checkmark    &              &&       \checkmark &  \checkmark & $\mathbb{T,P,B}$       \\
\citet{lopes2016simple}                &2016                 & Hybrid genetic algorithm       &   Integrated              &          &   \checkmark               &&          \checkmark     &             \checkmark &      $\mathbb{T,P,B}$       \\
\citet{quintero2017biased}           &2017                   & Biased randomized     &         Hierarchical                 &     \checkmark         &               &&        \checkmark       &    \checkmark          &        $\mathbb{P,B}$     \\
\citet{schneider2019large}     &2019                         & Tree-based search algorithm      & Hierarchical               &   \checkmark         &               &&              \checkmark &         \checkmark     &         $\mathbb{T,P,B,S}$    \\
\citet{accorsi2020hybrid}     &2020                          & AVXS     &Integrated                                       &      \checkmark        &               &&    \checkmark           &              $\times$&        $\mathbb{T}$     \\
\citet{akpunar2021hybrid}         &2021                      & Hybrid ALNS    &Integrated                                 &     \checkmark         &               &&   \checkmark            &      \checkmark        &         $\mathbb{T,P,B}$    \\
\citet{arnold2021progressive}      &2021                     & Progressive filtering           &   Hierarchical             &   \checkmark           &               &&     \checkmark          &        \checkmark      &      $\mathbb{T,P,B,S}$       \\
\citet{voigt2022hybrid}                     &2022            & HALNS             &      Integrated                        &            \checkmark  &               &&              \checkmark &             \checkmark &             $\mathbb{T,P,B}$      \\
This paper &-& Hybrid genetic algorithm&Mixed & &\checkmark &&\checkmark & \checkmark & $\mathbb{T,P,B,S}$\\
\hline
\end{tabular}
\end{center}
\end{tiny}
\end{table}

\subsection{Hierarchical approach}

The \textit{hierarchical approach} decomposes the CLRP into two subproblems, the facility location problem (FLP) and the MDVRP, which are solved separately. This approach has the advantage of allowing a direct application of many effective routing algorithms to solve the MDVRP with a fixed depot configuration. However, how to identify the best or most promising depot configurations remains a challenge, although some advanced strategies such as the progressive filtering heuristic \citep{arnold2021progressive} have been proposed.

\citet{tuzun1999two} introduced a pioneering hierarchical algorithm, the two-phase tabu search, for the CLRP. This algorithm integrates decision-making at two levels by using a two-phase tabu search: the first to determine the depot configuration and the second to optimize routes for the chosen depot configuration. These phases are carefully coordinated to efficiently explore the solution space. Experiments on 36 instances showed significant performance improvements over reference methods.


\citet{schneider2019large} proposed a tree-based search algorithm that systematically explores depot configurations in a tree-like fashion. The algorithm alternates between a location phase and a routing phase. First, it uses the minimum spanning forest to estimate the routing cost. Then, in the routing phase, it uses a special tabu search algorithm to improve the MDVRP solution. The algorithm matched or improved on the vast majority of previous best-known solutions for the instances of the  three classical sets, and demonstrated the ability to efficiently solve newly generated large-scale instances.

\citet{arnold2021progressive} developed a progressive filtering heuristic for the CLRP. The algorithm begins by empirically estimating an upper bound on the depots of the CLRP and employs a heuristic construction procedure to significantly reduce the number of promising configurations. Then, a dedicated MDVRP heuristic is applied to evaluate each of these promising depot configurations. Experimental results showed that the algorithm outperformed existing heuristics, especially on the largest benchmark instances.


\subsection{Integrated approach}

The \textit{integrated approach} considers the location decision and the routing problem simultaneously. Unlike the \textit{hierarchical approach}, the integrated approach manipulates different depot configurations by removing, opening, or swapping depots, whereas improving the routes cost by adjusting customers route assignments.


\citet{prins2006memetic} introduced a memetic algorithm with population management for the CLRP. The algorithm  employs a two-part chromosome encoding to represent each solution and uses the classical one-point crossover to generate offspring solutions whereas simultaneously reassigning the sequence of customers. For each offspring solution, the depot configuration and routes are constructed at the same time. This algorithm showed promising results on the  instances tested. 


\citet{lopes2016simple} introduced an evolutionary algorithm for the CLRP. The algorithm uses a route copy crossover to generate offspring solutions, allowing the discovery of new depot configurations during the crossover, and applies a local search procedure to improve the routing. The algorithm reported competitive results on the tested benchmark instances.

\citet{voigt2022hybrid} proposed a novel hybrid adaptive large neighborhood search algorithm (HALNS). The approach uses a population of solutions generated via an efficient adaptive large neighborhood search algorithm (ALNS), similar to \citep{hemmelmayr2012adaptive}. Notably, it combines two solutions by using ALNS to preserve selected parts of those solutions, rather than by using a traditional crossover operator. The algorithm explores depot configurations by removing, opening, and swapping depots during the destroy/repair process. The algorithm performed very well by finding three new best solutions.

As the review shows, promising depot configurations play an important role in both the hierarchical and integrated approaches, with a more pronounced role in the hierarchical approach. Meanwhile, population-based algorithms \citep{prins2006solving,lopes2016simple} have achieved encouraging performances. In this work, we revisit the integrated approach by proposing an innovative population-based algorithm. The algorithm organizes the solutions of the population into subpopulations according to the depot configurations, and uses a special crossover to generate promising offspring solutions by mixing different depot configurations and assembling route edges. The algorithm additionally integrates an advanced population management to maintain a pool of high-quality and diverse solutions, a coverage ratio heuristic to generate initial depot configurations, and routing heuristics for effective routing optimization.


\section{General approach of HGAMP}\label{memetic_algo}

\begin{figure}[H]
\includegraphics[width=5.0in]{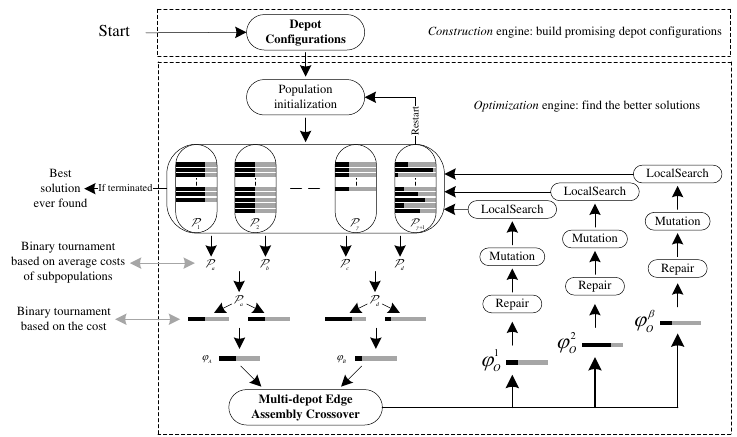}
\caption{Flow chart of HGAMP for the CLRP.}
\label{flow_chart}
\end{figure}

The proposed HGAMP algorithm includes a \textit{construction} engine (Section \ref{crh_algo}) to build a set of promising depot configurations, and an \textit{optimization} engine (Section \ref{Optimization_engine})  to improve the solutions. As illustrated in Figure \ref{flow_chart} and Algorithm \ref{algo_hga}, the \textit{construction} engine uses a coverage ratio heuristic algorithm (CRH) to identify promising depot configurations. For each selected depot configuration $\mathcal{D}_i$, a subpopulation $\mathcal{P}_i$ is initialized with multiple solutions generated using a random greedy heuristic. Subsequently, the \textit{optimization} engine evolves the population $\mathcal{P}$ over a number of generations, by utilizing crossover, repair, mutation, local search, and population management. Each offspring solution is assigned to the subpopulation corresponding to its depot configuration and is managed by that subpopulation. HGAMP stops and returns the best solution $\varphi^*$ when the predefined stopping condition is met (e.g., a maximum cutoff time or a maximum number of iterations). 

\renewcommand{\baselinestretch}{0.8}\huge\normalsize
\begin{algorithm}[H]\label{algo_hga}
\begin{footnotesize}
\caption{Main framework of the HAGMP algorithm}
\DontPrintSemicolon 
\KwIn{Instance $I$;}
\KwOut{The best solution $\varphi^{*}$ found so far;}
\Begin{
$\mathcal{D}=\{\mathcal{D}_1,\mathcal{D}_2,\cdots,\mathcal{D}_\gamma\} \gets CRH(I)$;\tcc*[l]{Identify a set of promising depot configurations, Appendix A of the online supplement}
$\mathcal{P}=\{\mathcal{P}_1,\mathcal{P}_2,\cdots,\mathcal{P}_\gamma,\mathcal{P}_{\gamma+1}\} \gets PopInitial(\mathcal{D}, I)$ \tcc*[l]{Initialize population $\mathcal{P}$, Section \ref{initial_solution}}
$\varphi^{*} \gets \arg\min \{f(\varphi)| \forall \varphi \in \mathcal{P}\}$;\tcc*[l]{Record the best solution found so far}
\While{$Stopping\ condition\ is\ not\ met$}{
	$\{\varphi_A,\varphi_B\} \gets $\textit{ParentSelection}($\mathcal{P}$);\tcc*[l]{Select two parent solutions}
	$\{\varphi^1_O,\varphi^2_O,\cdots,\varphi^\beta_O\} \gets $\textit{mdEAX}($\varphi_A$, $\varphi_B$);\tcc*[l]{Generate offspring solutions, Section \ref{EAX_section}}
	\For{$i=1$ to $\beta$}{
		$\varphi^i_O \gets$ \textit{Repair}($\varphi^i_O$);\tcc*[l]{Restore feasibility, Section \ref{repair_section}}
		$\varphi^i_O \gets$ \textit{Mutation}($\varphi^i_O$);\tcc*[l]{Generate mutation, Section \ref{mutation_section}}
		$\varphi^i_O \gets$ \textit{LocalSearch}($\varphi^i_O$);\tcc*[l]{Improve the offspring solution, Section \ref{ls_section}}
		 \If {$f(\varphi^i_O) < f(\varphi^{*})$} {
	    	$\varphi^{*} \gets \varphi^i_O$;\\
    	}
		$\{\mathcal{P}, \mathcal{D}\} \gets \mathit{ManagingPop(\mathcal{P},\mathcal{D},\varphi^i_O, \varphi^*)}$;\tcc*[l]{Manage the population, Section \ref{pop_man_section} }	
	}
}
\Return{$\varphi^{*}$};\\
}
\end{footnotesize}
\end{algorithm}
\renewcommand{\baselinestretch}{1.0}\huge\normalsize

HGAMP incorporates three particularly interesting innovations. First, its multi-population scheme effectively ensures a high degree of diversity among the population solutions, which helps to reduce the risk of premature convergence of the algorithm. Second, its mdEAX crossover has the ability to generate new promising solutions by creating new depot configurations whereas inheriting route edges from the parent solutions. Third, its CRH heuristic contributes to the identification of promising depot configurations, which is helpful for a better convergence of the algorithm.

\subsection{Construction engine}\label{crh_algo}


We use a general coverage ratio heuristic (CRH) (see Appendix A of the online supplement) composed of two filtering techniques to identify promising depot configurations.

The preliminary filtering identifies a set of depot configurations in two steps, considering a balance between estimated route costs and geographic dispersion of the candidate depots. The first step uses a minimum spanning tree (MST) algorithm to estimate approximate routing costs for each depot intended for integration into the final depot configuration. The second step progressively selects the candidate depots, taking into account both the approximate costs (consisting of a fixed cost $o_i$ and estimated routing costs) and the geographic dispersion of the candidate depots. To measure the geographic dispersion, we introduce an overlap coverage ratio. Specifically, each depot $i$ covers a set of customers $\mathcal{S}_i$, determined by the MST algorithm. The ratio for depot $i$ is defined as $r_i = |\mathcal{S}_i \cap \mathcal{S}_{\mathcal{H}}|/{|\mathcal{S}_i|}$, where $\mathcal{S}_{\mathcal{H}}$ represents the set of customers covered by a selection of depots $\mathcal{H}$. Therefore, a smaller value of $r_i$ signifies a more favorable geographic dispersion for depot $i$. For each candidate depot, once the overlap coverage ratio satisfies a dynamically adjusted threshold, it is placed in a sorted list based on its costs and regarded as a candidate item. To select a candidate depot from this list, a greedy random method is employed. Notably, the threshold is dynamic in nature, as the overlap coverage ratio tends to increase with the inclusion of more depots, covering a greater number of customers. The process iterates until a feasible depot configuration is achieved, ensuring that the cumulative capacity, denoted as $\sum_{i\in \mathcal{H}}w_i$, surpasses the total demand of customers. Through these two sequential steps, a set of $H_{max}$ distinct depot configurations are generated. 

From these preliminarily selected depot configurations, we use the secondary filter to determine a reduced set of final depot configurations. For each candidate depot configuration $\mathcal{D}_i$, we use it to generate a limited number $N_t$ (a parameter, fixed to 10) of CLRP solutions, which are immediately improved by the local search procedure of Section \ref{ls_section}. We then calculate the average cost of these $N_t$ improved solutions, which is used to assess the attractiveness of the depot configuration $\mathcal{D}_i$. Based on the attractiveness values of the $H_{max}$ depot configurations, the first $\gamma$ top depot configurations are identified as the final promising depot configurations, which are used by the search algorithm. 


\subsection{Optimization engine}\label{Optimization_engine}

From the $\gamma$ promising depot configurations $\mathcal{D}$, a population $\mathcal{P}$, which is organized into $\gamma$ subpopulations, is constructed and then evolved using crossover, mutation, local search and population management.

\subsubsection{Population initialization}\label{initial_solution}


Let $\mathcal{D}=\{\mathcal{D}_1,\mathcal{D}_2,\dots,\mathcal{D}_{\gamma}\}$ be the set of $\gamma$ promising depot configurations from the construction engine. For each depot configuration $\mathcal{D}_i$, a subpopulation $\mathcal{P}_{i}$ is created whose solutions are all based on this depot configuration. An additional subpopulation $\mathcal{P}_{\gamma+1}$ is also built to promote and maintain diversity, containing solutions that are not related to any of the $\gamma$ depot configurations.

To initialize the subpopulation $\mathcal{P}_{i}$ for a given depot configuration $\mathcal{D}_i$, a random greedy heuristic (RGH) is used to construct each solution, which is then improved by local search to increase its quality. RGH constructs an initial solution using the following steps. First, it randomly selects a  depot from the given depot configuration and  uses the cheapest customer to start a route. Second, it greedily inserts other unvisited customers into the route based on the nearest neighbors rule (Section \ref{ls_section}). When the capacity of the vehicle is reached, RGH starts a new route. When the maximum capacity of the current depot is exceeded, RGH selects a new depot from the given depot configuration. RGH continues this process until all customers are covered, resulting in a feasible initial solution. 

Each initial solution is further improved by the local search procedure of Section \ref{ls_section} and then added to the subpopulation. When $\mathcal{P}_i$ reaches the maximum limit of $\mu + \lambda$ ($\mu$ and $\lambda$ are two parameters, see Section \ref{pop_man_section}), the advanced updating strategy (ADQ, described in Section \ref{pop_man_section}) is triggered to retain $\mu$ individuals. This procedure terminates and returns subpopulation $\mathcal{P}_i$ once $4 \times \mu$ solutions have been generated.

To construct the last subpopulation $\mathcal{P}_{\gamma+1}$, which contains solutions with no fixed depot configurations, a random greedy technique is used to select the depots for each initial solution based on their corresponding rough costs $u_i$, introduced in Section A.1 of the online supplement. At the beginning, all candidate depots and their rough costs are sorted in a list, denoted by $\mathcal{L}$. Each depot is then selected with a certain probability (see Algorithm 1 of the online supplement). Once the total capacity of the selected depots covers the total demand of the customers, a complete depot configuration is obtained. Unlike CRH, this method does not take into account geographic dispersion of the candidate depots. After considering $4 \times \mu$ solutions, subpopulation $\mathcal{P}_{\gamma+1}$ is successfully constructed.

\subsubsection{Multi-depot edge assembly crossover}\label{EAX_section}


Prior to the application of the crossover operator, two parent solutions from two subpopulations are selected in two steps. First, two subpopulations are selected using a binary tournament strategy based on {their average objective values $\widehat{\mathcal{P}_i} = \sum_{\varphi_j\in \mathcal{P}_i}f(\varphi_j)/{|\mathcal{P}_i|}$, where $f(\varphi)$ is the objective of solution $\varphi$. Second, another binary tournament selection is applied within each subpopulation to select a parent based on its objective value (see Figure \ref{flow_chart}).

From the parent solutions that are necessarily based on two different depot configurations, the proposed mdEAX crossover combines the parents to generate offspring solutions by considering their depot configurations as well as their associate routes. Like the crossovers in \citet{nagata2009edge} for the VRP, \citet{he2022general} for the split delivery VRP, and \citet{he2023hybrid} for the TSPs with profits, our mdEAX is based on the general idea of the popular EAX crossover for the TSP \citep{nagata2013powerful}. However, unlike these studies, mdEAX is designed to be able to consider routing solutions with different depot configurations, which is not the case in the previous studies.

Given two parent solutions $\varphi_A$ and $\varphi_B$, we build two graphs $\mathcal{G_A} = (\mathcal{V_A, E_A})$ and $\mathcal{G_B} = (\mathcal{V_B, E_B})$ where $\mathcal{V_A}$ and $\mathcal{V_B}$ are the sets of vertices representing the depots and customers visited by $\varphi_A$ and $\varphi_B$, respectively, and $\mathcal{E_A}$ and $\mathcal{E_B}$ are the sets of edges traversed by $\varphi_A$ and $\varphi_B$, respectively. Because each customer is visited exactly once, each customer vertex in $\mathcal{G_A}$ and $\mathcal{G_B}$ has a degree of two. However, the vertices associated to the depots of $\varphi_A$ and $\varphi_B$ may have different degrees because the parents use two different depot configurations (e.g., the two red depots in Figure \ref{fig_eax}, unopened in $\varphi_A$ with a degree of zero, and opened in $\varphi_B$ with degrees of 4 and 2). Following \citep{he2022general,he2023hybrid}, we extend $\mathcal{E_A}$ and $\mathcal{E_B}$ by adding dummy loops to make each vertex in $\mathcal{G_A}$ have the same degree as in $\mathcal{G_B}$, resulting in the extended edge sets $\mathcal{E}^{'}_\mathcal{A}$ and $\mathcal{E}^{'}_\mathcal{B}$. Then, the joint graph $\mathcal{G}_\mathcal{AB} = (\mathcal{V_A}\cup \mathcal{V_B}, (\mathcal{E}^{'}_\mathcal{A} \cup \mathcal{E}^{'}_\mathcal{B})\backslash (\mathcal{E}^{'}_\mathcal{A} \cap \mathcal{E}^{'}_\mathcal{B}))$ is created and used by the mdEAX crossover. 

Specifically, from parent solutions $\varphi_A$ and $\varphi_B$, the mdEAX crossover generates $\beta$ offspring solutions ($\beta$ is a prefixed threshold value) by the following steps.



\begin{figure}[htbp]
\centering
\includegraphics[width=6.3in]{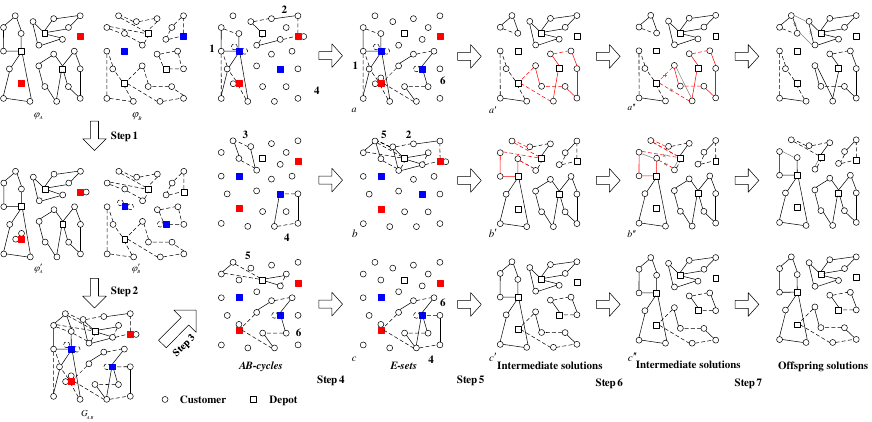}
\centering
\caption{Illustration of the mdEAX crossover for the CLRP}
\label{fig_eax}
\end{figure}



\begin{enumerate}
\item Add dummy loops. Dummy loops are introduced to extend graphs $\mathcal{G_A}$ and $\mathcal{G_B}$ to ensure that the \textit{degree difference} becomes 0 for each depot. As illustrated in Figure \ref{fig_eax}, multiple dummy loops are added to the colored depots of $\varphi_A$ and $\varphi_B$ to make sure that each depot vertex has the same degree in the extended graphs $\mathcal{G}^{'}_\mathcal{A}=(\mathcal{V_A}, \mathcal{E}^{'}_\mathcal{A})$ and $\mathcal{G}^{'}_\mathcal{B}=(\mathcal{V_B}, \mathcal{E}^{'}_\mathcal{B})$.

\item Build the joint graph  $\mathcal{G}_\mathcal{AB} = (\mathcal{V_A}\cup\mathcal{V_B}, (\mathcal{E}^{'}_\mathcal{A} \cup \mathcal{E}^{'}_\mathcal{B})\backslash (\mathcal{E}^{'}_\mathcal{A} \cap \mathcal{E}^{'}_\mathcal{B}))$ from $\mathcal{G}^{'}_\mathcal{A}$ and $\mathcal{G}^{'}_\mathcal{B}$.

\item Construct \textit{AB-cycles} from $\mathcal{G_{AB}}$. The edges in $\mathcal{G_{AB}}$ are partitioned into disjoint \textit{AB-cycles} using the method described in \cite{he2022general}. An AB-cycle is a cycle such that its edges alternates between edges of $\mathcal{E}^{'}_\mathcal{A}$ and  edges of  $\mathcal{E}^{'}_\mathcal{A}$. Because each vertex in $\mathcal{G_{AB}}$ is connected to an even number of edges, and half of these edges come from $\mathcal{E}^{'}_\mathcal{A}$ whereas the other half come from $\mathcal{E}^{'}_\mathcal{B}$, the edges in $\mathcal{G_{AB}}$ can be partitioned into \textit{AB-cycles} accurately. In Figure \ref{fig_eax}, the edges in $\mathcal{G_{AB}}$ are partitioned into six disjoint \textit{AB-cycles}.

\item Generate \textit{E-sets}. 
To ensure a sufficient diversity in the offspring solutions, it is essential to combine \textit{AB-cycles} that share common vertices into larger \textit{E-set}. This is because if the size of \textit{AB-cycles} is too small, the resulting offspring solutions could be too similar to the parent solutions. After constructing \textit{AB-cycles}, they are grouped into \textit{E-sets}. If the number of \textit{E-sets} exceeds the threshold $\beta$, some \textit{E-sets} are randomly merged to retain only $\beta$ \textit{E-sets}. In Figure \ref{fig_eax}, three \textit{E-sets} are generated from the six \textit{AB-cycles}. 

\item Construct intermediate solutions. A random solution $\varphi$ is selected first, which can be $\varphi_A$ or $\varphi_B$. For each \textit{E-set} $\mathcal{E}_s$, we create an intermediate solution $\varphi' = (\mathcal{E_A}\backslash(\mathcal{E}_s \cap \mathcal{E_A})) \cup (\mathcal{E}_s \cap \mathcal{E_B})$, where the added dummy loops are ignored.

\item Split mega tours. An intermediate solution may contain routes visiting more than one depot (we call them mega tours), as we observe in solution $a'$ (red lines) of Figure \ref{fig_eax}. For these tours, we use a new greedy strategy to remove extra depots by taking into account the capacity of opening depots. Precisely, given a tour that visits two depots $i_1$ and $i_2$ with a residual capacity (which may be negative) $r_{i_1}$ and $r_{i_2}$, respectively, we remove depot $i_1$ if $r_{i_1} < r_{i_2}$; otherwise, we remove depot $i_2$. If no capacity constraints are imposed on depots, the worst depot with respect to the traveling cost is removed. For example, the tour in $a'$ is split by removing two edges and adding an edge, as is also done for solution $b'$. See the repaired feasible intermediate solutions after Step 6 in Figure \ref{fig_eax}. If the mega tour has more than two depots, we repeat this process until only one depot remains.

\item Eliminate subtours. Some intermediate solutions may contain some isolated tours (called subtours), such as the triangle subtour in the upper-left corner of solution $a{''}$ in Figure \ref{fig_eax}. The 2-opt* operator \citep{potvin1995exchange} is applied to link these isolated subtours to other tours. 

\end{enumerate}

Because the complexity of splitting mega tours is limited to $\mathcal{O}(m)$, the time complexity of mdEAX is $\mathcal{O}{((n+m) \times \alpha)}$, as other steps of mdEAX are the same as gEAX in \cite{he2022general}. Here, $\alpha$ is the number of nearest neighbors (see Section \ref{ls_section}). Furthermore, the space complexity is bounded by $\mathcal{O}(|\mathcal{E}^{'}_\mathcal{A}|)$, assuming that $|\mathcal{E}^{'}_\mathcal{A}| \geq |\mathcal{E}^{'}_\mathcal{B}|$.

The mdEAX crossover not only extracts backbone information from parent solutions and transmits it to descendants, but also effectively promotes diversity. In addition, it can generate new promising depot configurations when building descendant solutions. As such, it plays a key role in our integrated approach, which simultaneously considers depot decision and route optimization, and avoids the difficulty of the hierarchical approach, which requires the prior determination of promising depot configurations.

\subsubsection{Restore the feasibility of offspring individuals}\label{repair_section}

The mdEAX crossover ignores the capacity constraints. Therefore, an offspring solution may violate these constraints. To address this issue, we propose a greedy procedure that utilizes a generalized cost function to eliminate all violations of both depot and vehicle capacities. 

Let $\varphi$ be an offspring solution, let $T_\varphi$ denote the cumulative depot capacity for the given depot configuration $\mathcal{D}_\varphi$ and $T$ denote the total demand of customers. Then $T_\varphi< T$ or $T_\varphi \geq T$. In the first case, we need to select additional depots to expand $\mathcal{D}_\varphi$. A customer is removed from the route that violates the depot capacity and inserted into a new route with a new open depot with respect to traveling costs, whereas ignoring the fixed depot and vehicle costs to improve feasibility. This process continues until $T_\varphi \geq T$. Then, we restore the capacity constraints for both vehicles and depots. Four well-known inter-route neighborhood operators (Relocate and Swap with single customers and 2-Opt* in Section \ref{ls_section}) are used to explore the neighborhood of solutions based on Eq. (\ref{eq1}), where $f(\varphi)$ denotes the objective function value of solution $\varphi$, $f_l(\varphi)$ represents the cumulative overcapacity of depots, $f_r(\varphi)$ is the cumulative overcapacity of routes, and $p_l$ and $p_r$ are penalty factors for overcapacity of depots and routes, respectively. To avoid excessive iterations, we use a tabu list to prevent reversing a performed move. If a feasible move cannot be found whereas the solution is still infeasible, we multiply the penalty factors $p_l$ and $p_r$ by 10. The procedure continues until the solution becomes feasible. We set $p_l$ and $p_r$ to 1 initially. 
\begin{equation}\label{eq1}
f_g(\varphi) = f(\varphi) + p_l \times f_l(\varphi) + p_r \times f_r(\varphi)
\end{equation}

\subsubsection{Mutation}\label{mutation_section}

In HGAMP, the population $\mathcal{P}$ faces a challenging problem where the edges of the offspring solutions are almost exclusively from their parents, leaving little room for new edges in the solutions newly added to the population. To introduce enough diversity into the offspring solutions, we apply a mutation operator, based on an idea presented in \citet{shaw1998using}, to modify each offspring solution with probability $\zeta$. Specifically, the operator removes a number of customers that are similar with respect to a predefined characteristic (e.g. location or demand). The similarity between two customers is calculated based on their distance. At the beginning, the first customer is randomly selected, and then other customers are selected by using the roulette-wheel selection with respect to similarity. After selecting $\xi$ customers, all of these customers are removed from the solution. Then, we greedily insert the removed customers back into the partial solution.


\subsubsection{Local search}\label{ls_section}


For local optimization, HGAMP adopts ten commonly used neighborhood operators from \citet{schneider2019large,duhamel2010grasp} and explores them under the framework of variable neighborhood descent.These neighborhood operators include Relocate and Swap generalized to sequences of two consecutive customers, as well as 2-Opt and 2-Opt*. The neighborhoods are restricted to those involving geographically close customer pairs (\textit{i},\textit{j}), where customer \textit{j} belongs to the $\alpha$-nearest neighbor from customer \textit{i}. The granularity parameter $\alpha$ limits the time complexity of all neighborhood operators to $\mathcal{O}((n+m)\times\alpha)$.

\subsubsection{Population management}\label{pop_man_section}

The primary objective of population management is to maintain a healthy diversity of the population $\mathcal{P}$ throughout the search process. This involves two key tasks: 1) updating the depot configurations when a better one is discovered, and 2) managing and updating each subpopulation. 

Depot configurations $\mathcal{D}$ is updated each time the global best solution $\varphi^*$ is replaced by the current solution $\varphi$ whose corresponding depot configuration $\mathcal{D}_\varphi$ is not in set $\mathcal{D}$ ($\mathcal{D}_{\varphi}\notin\mathcal{D}$). The worst subpopulation $\mathcal{P}_w$ is identified among the first $\gamma$ subpopulations and its corresponding individuals are discarded. The depot configuration $\mathcal{D}_w$ is then replaced with the new depot configuration $\mathcal{D}_\varphi$, and subpopulation $\mathcal{P}_w$ is re-initialized based on $\mathcal{D}_\varphi$. 

Furthermore, an advanced distance-and-quality updating rule (ADQ) \citep{he2022general,vidal2012hybrid} is used to manage each subpopulation. Each new offspring solution is inserted into the corresponding subpopulation based on its depot configuration, and clones are not permitted. For each subpopulation, when the number of solutions reaches the maximum size $\mu + \lambda$ ($\lambda$ is the generation size), $\lambda$ solutions are removed based on a biased fitness, and only $\mu$ solutions go to the next generation. We use the method proposed in \cite{prins2006solving} to compute the Hamming distance between two solutions. 

If the best solution found so far $\varphi^*$ cannot be improved for $\eta$ consecutive invocations of the local search ($\eta$ is a parameter called the population rebuilding threshold), the algorithm restarts by generating a completely new population. In fact, a large number of subpopulations will slow down the convergence of the algorithm. So, when the algorithm reaches half the total time budget, we reduce the number of subpopulations by half, keeping only the most promising subpopulations with respect to the average objective values $\hat{\mathcal{P}_i}$.  

\subsection{Discussion}\label{Discussion}

HGAMP enhances the canonical hybrid genetic algorithm framework \citep{he2022general,vidal2012hybrid} with three vital techniques: the multi-population scheme, the mdEAX crossover, and the coverage ratio heuristic CRH. First, the multi-population scheme allows the algorithm to manage multiple depot configurations and their corresponding solutions. This scheme provides the search algorithm with a better coverage of the search space, increasing its chances of finding high-quality solutions. Second, to generates offspring solutions and thanks to the multi-population scheme, the mdEAX crossover recombines parent solutions with different depot configurations, taking into account the selection of depots and the assembly of route edges. Finally, the CRH algorithm is proposed to identify a number of initial promising depot configurations considering the best balance between cost efficiency and geographic distribution.

Beyond the CLRP studied in this paper, the proposed approach, especially, the above three key components, can be advantageously adapted to solve related LRPs such as the 2E-VRP \citep{perboli2011two} and the location arc routing problem \citep{lopes2014location}. The generality of our approach to simultaneously explore unknown depot configurations and new route solutions by edge assembly can serve as a relevant solution approach to bridge the gaps encountered in the context of diverse LRPs. Finally, the CRH heuristic provides a unified and alternative way to identify promising depot configurations for LRPs.

\section{Experimental Evaluation and Comparisons}\label{com_results}

The purpose of this section is to evaluate the performance of HGAMP through experiments and to compare its results with those of state-of-the-art algorithms.

\subsection{Benchmark instances}\label{instances}


There are three classical benchmark sets that include 79 small and medium instances from \citet{tuzun1999two}, \citet{prins2006solving} and \citet{barreto2007using}. These sets have been widely tested in the literature, and their results are difficult to  further improve upon. These sets are denoted by $\mathbb{B}, \mathbb{P}, \mathbb{T}$. In addition, \citet{schneider2019large} introduced a rich and challenging set (denoted by $\mathbb{S}$) of 202 instances with different characteristics and sizes. These 281 instances and the best solutions obtained by HGAMP are available online (https://github.com/pengfeihe-angers/CLRP.git).

\begin{itemize}
\item Set $\mathbb{B}$: This set includes 13 instances (7 with a proven optimal value) with 21 to 150 customers and 5-14 depots, and in most instances a depot capacity is imposed. 

\item Set $\mathbb{P}$: This set contains 30 instances (20 with a proven optimal value) with 20 to 200 customers and 5 to 10 capacitated depots. 

\item Set $\mathbb{T}$: This set includes 36 instances (6 with a proven optimal value)  with 100 to 200 customers and 10-20 uncapacitated depots. 

\item Set $\mathbb{S}$: This set includes 202 instances (64 with a proven optimal value) with varying characteristics, such as the number of customers, depots, vehicle capacity, and depot costs. The instances have 100 to 600 customers and 5-30 capacitated depots. The instances are named in the format of $n-m-tz$, where $n$ represents the number of customers, $m$ is the number of depots, $t$ indicates the geographic distribution, and $z$ denotes the vehicle capacity and depot costs. 
\end{itemize}


\subsection{Experimental protocol and reference algorithms}\label{exper_prococol}

\textbf{Parameter setting.} The HGAMP algorithm has six parameters: the minimum size of each subpopulation $\mu$, the generation size $\lambda$, the granularity threshold of nearest neighbors $\alpha$, the mutation probability $\zeta$, the mutation length $\xi$ and the population rebuilding threshold $\eta$. To calibrate these parameters, the automatic parameter tuning package Irace \citep{lopez2016irace} is used. The tuning was performed on 10 instances with 100-300 customers. The tuning budget was set to be 2000 runs. Table \ref{table_parameter} shows the candidate values for the parameters and the final values suggested by Irace. This parameter setting can be considered the default setting of HGAMP and was used consistently in our experiments. 

\begin{table}[h]
\caption{Parameter tuning results.} \label{table_parameter}
\renewcommand{\baselinestretch}{}\huge\normalsize
\begin{scriptsize}
\begin{tabular}{p{1.5cm}p{1.1cm}p{5.0cm}p{4.5cm}p{2.0cm}}
\hline
Parameter & Section & Description & Candidate values & Final values\\
\hline
$\mu$						&\ref{pop_man_section}  & minimal size of each subpopulation &$\{10,20,30,40,50\}$  &30\\
$\lambda$						& \ref{pop_man_section} & generation size & $\{10,20,30,40,50\}$ &30\\
$\alpha$							&\ref{ls_section}          & granularity threshold & $\{5,10,15,20,25,30\}$ &20 \\
$\zeta$								&\ref{mutation_section} & mutation probability & $\{0,0.05,0.1,0.15,0.2,0.25,0.3\}$  &0.15\\
$\xi$									&\ref{mutation_section}& mutation length & $\{0.05,0.1,0.15,0.2,0.25\}$ &0.25\\
$\eta$                       &\ref{pop_man_section}&population rebuilding threshold &$\{30000,50000,70000,90000\}$& 70000\\
\hline
\end{tabular}
\end{scriptsize}
\end{table}

\textbf{Reference algorithms.}
We adopt the following best CLRP heuristic algorithms as well as the best known solutions BKS (best upper bounds) reported in the literature as the references for comparative experiments. 

\begin{itemize}

\item BKS. This indicates the best known solutions (upper bounds) that are summarized from the state-of-the-art heuristic algorithms \citep{arnold2021progressive,voigt2022hybrid,schneider2019large,contardo2014grasp,lopes2016simple}, and the cutting-edge exact approaches \citep{liguori2022non,contardo2014exact,baldacci2011exact}.

\item AVXS \citep{accorsi2020hybrid}. The algorithm was coded in C++ and executed on a computer with an Intel i7-8700k CPU at 3.7 GHz and 32 GB RAM.  One notices that the algorithm only reported excellent results on set $\mathbb{T}$. 

\item HALNS \citep{voigt2022hybrid}. The hybrid adaptive large neighborhood search was implemented in C++, running on a computer with an AMD Ryzen 9 3900X CPU at 3.8 GHz and 32 GB RAM. Notably, the algorithm reported excellent results on the three classical sets $\mathbb{B}$, $\mathbb{P}$ and $\mathbb{T}$. 

\item PF \citep{arnold2021progressive}. This algorithm was coded in Java and the experiments were conducted on a computer with an Intel i7 4790 at 3.6 GHz and 8 GB RAM. One notes that parallel computing techniques were employed to reduce running time. The algorithm reported excellent results on the four sets of benchmark instances. 

\item TBSA$_{basic}$  \citep{schneider2019large}. The algorithm, implemented in C++ on a computer with Xeon E5-2670 at 2.6 GHz and 32 GB RAM, reported excellent results on the four sets of benchmark instances. 

\item TBSA$_{quality}$ \citep{schneider2019large}. The algorithm was implemented in C++, running on a computer with Xeon E5-2670 at 2.6 GHz and 32 GB RAM, only reported excellent results on three classical sets $\mathbb{B}$, $\mathbb{P}$ and $\mathbb{T}$. One notices that it is identical to TBSA$_{basic}$ except for an increased number of iterations in the location phase.

\end{itemize}

\subsection{Variants of HGAMP}\label{var_hgamp}

HGAMP uses its CRH heuristics to identify a set of initial depot configurations $\mathcal{D}$. In addition to the CRH method, we also explored the progressive filtering method proposed by \citet{arnold2021progressive} to generate initial depot configurations. For an instance, we ran the progressive filtering and saved the depot configurations of its first round in $\mathcal{D}_{ps}$ ($0 \leq |\mathcal{D}_{ps}|\leq 100$). Based on this, we obtain the following three algorithms, where $\mathcal{D}$ is the set of initial depot configurations that will be used by HGAMP to start its search.

\begin{itemize}
\item  HGAMP$_b$: this HGAMP algorithm uses CRH to generate initial depot configurations.
\item HGAMP$_p$: this variant uses the depot configurations from the progressive filtering (i.e., $\mathcal{D} \gets \mathcal{D}_{ps}$).
\item HGAMP$_m$: this variant uses the preliminary filter of CRH to generate initial depot configurations $\mathcal{D}\gets$ Algorithm 1 of the online supplement and then $\mathcal{D} \gets \mathcal{D} \cup \mathcal{D}_{ps}$.
\end{itemize}
 
\textbf{Experimental setting and stopping criterion.} 
The HGAMP algorithm was implemented in C++ and compiled using the g++ compiler with the -O3 option (The code of the HGAMP algorithm will be available at: https://github.com/pengfeihe-angers/CLRP.git). All experiments were conducted on an Intel Xeon E-2670 processor of 2.5 GHz and 2 GB RAM running Linux with a single thread. The HGAMP algorithm terminates after a maximum of 300,000 iterations, where one iteration means that one offspring solution is constructed and improved by the local search subsequently. Additionally, we use HGAMP$_b$-{15w} and HGAMP$_b$-45w to represent the HGAMP$_b$ algorithm, which terminates after a maximum of 150,000 and 450,000 iterations, respectively. Whereas the reference algorithms use different stopping conditions, the time scale used for computation remains comparable. To facilitate a meaningful comparison of HGAMP's computational efficiency, we have adjusted HGAMP's stopping condition to the time similar to that of TBSA \citep{schneider2019large} and HALNS \citep{voigt2022hybrid}.

\subsection{Computational results and comparisons}\label{comparison1}
In the following subsections, we present HGAMP's results on the benchmark instances and compare them to the reference algorithms. 

\subsubsection{Performance analysis on the three classical sets}\label{class_results}
Table \ref{sumResults_threeSet} summarizes the results obtained by the HGAMP algorithm on the three classical sets, comparing them to the reference algorithms in terms of both best and average values. For each benchmark set, we show the number of instances where our algorithm (HGAMP) has a better (Win), equal (Tier) or worse (Loss) result compared to each reference algorithm, in terms of the best and average results. Wilcoxon signed-rank $p$-values are also included. The '-' symbol indicates that the result is unavailable. Tables 1-2 of the online supplement provide detailed results for these 79 instances. Additionally, Figure \ref{perfor_classical} illustrates the performance comparison over the sets $\mathbb{P}$ and $\mathbb{T}$ with the running times of all methods scaled with the Passmark scores (https://www.cpubenchmark.net/singleThread.html\#).

For the 13 instances of set $\mathbb{B}$, HGAMP$_b$ achieves the same performance as BKS, HALNS, and TBSA$_{quality}$ in terms of the best results. Moreover, our algorithm has a slight advantage over the reference algorithms in terms of the average results. Because the instances in this set are rather small and state-of-the-art algorithms produce identical results, this set may not be challenging enough to evaluate the performance of new algorithms.

For the 30 instances of set $\mathbb{P}$, HGAMP$_b$ improves upon one BKS (indicated in boldface) and matches 23 BKS. Compared to the state-of-the-art reference algorithms, HGAMP$_b$ achieves results similar to HALNS, TBSA$_{basic}$, and TBSA$_{quality}$ with no statistical significances in terms of the best results. From Figure \ref{perfor_classical}(a), we notice that TBSA$_{basic}$ dominates all approaches for shorter running times, whereas TBSA$_{quality}$ still outperforms other algorithms for longer running times, though the difference are very marginal.

For the 36 instances of set $\mathbb{T}$, HGAMP$_b$ improves upon the BKS in two cases (indicated in boldface) and matches the BKS in 22 other cases. Our algorithm outperforms three reference algorithms AVXS, PF and TBSA$_{basic}$, in terms of the best results. Furthermore, as shown in Figure \ref{perfor_classical}, whereas HALNS dominates all algorithms for shorter running times, the differences in the average gap to BKS for most compared algorithms remain marginal.

\begin{table}[h]
\caption{Summary of results between the HGAMP$_b$ and reference algorithms on three classical sets of 79 instances.}\label{sumResults_threeSet}
\begin{scriptsize}
\begin{center}
\begin{tabular}{p{1.2cm}p{3.6cm}p{0.9cm}p{0.9cm}p{0.9cm}p{1.5cm}p{0.01cm}p{0.9cm}p{0.9cm}p{0.9cm}p{1.5cm}}
\hline
\multirow{2}{*}{Instances} & \multicolumn{1}{c}{\multirow{2}{*}{Pair algorithms}} & \multicolumn{4}{c}{Best}               &  & \multicolumn{4}{c}{Avg.}               \\
\cline{3-6} \cline{8-11}
  & \multicolumn{1}{c}{}   & \#Wins & \#Tiers & \#Losses & \textit{p}-value  &  & \#Wins & \#Tiers & \#Losses & \textit{p}-value  \\
\hline
\multirow{4}{*}{$\mathbb{B} (13)$}        & HGAMP$_b$ vs. BKS & 0 & 13 &0 &1.00E+00 && -&-&-&-\\
 							& HGAMP$_b$ vs. PF   & 2      & 11      & 0        & 5.69E-01 &  & -      & -       & -        & -        \\
                           & HGAMP$_b$ vs. HALNS                                   & 0      & 13      & 0        & 1.00E+00 &  & 1      & 11      & 1        & 3.30E-01 \\
                           & HGAMP$_b$ vs. TBSA$_{basic}$                             & 2      & 11      & 0        & 9.26E-01 &  & 4      & 8       & 1        & 5.88E-01 \\
                           & HGAMP$_b$ vs. TBSA$_{quality}$                           & 0      & 13      & 0        & 1.00E+00 &  & 3      & 9       & 1        & 8.50E-01 \\
\hline\multirow{5}{*}{$\mathbb{P} (30)$}         & HGAMP$_b$ vs. BKS                                     & \textbf{1}      & 23      & 6        & 4.69E-02 &  & -      & -       & -        & -        \\
                           & HGAMP$_b$ vs. PF                                      & 12     & 16      & 2        & 5.25E-03 &  & -      & -       & -        & -        \\
                           & HGAMP$_b$ vs. HALNS                                   & 3      & 24      & 4        & 6.88E-01 &  & 3      & 15      & 12       & 8.33E-02 \\
                           & HGAMP$_b$ vs. TBSA$_{basic}$                             & 5      & 20      & 5        & 9.22E-01 &  & 9      & 14      & 7        & 9.59E-01 \\
                           & HGAMP$_b$ vs. TBSA$_{quality}$                           & 3      & 22      & 5        & 3.13E-01 &  & 3      & 15      & 12       & 2.01E-03 \\
\hline

\multirow{6}{*}{$\mathbb{T} (36)$}         & HGAMP$_b$ vs. BKS       & \textbf{2}      & 22      & 12       & 3.59E-03 &  & -      & -       & -        & -        \\
                           & HGAMP$_b$ vs. AVXS   & 17     & 13      & 6        & 3.94E-03 &  & 23     & 5       & 8        & 5.19E-03 \\
                           & HGAMP$_b$ vs. PF   & 25     & 9       & 2        & 2.86E-05 &  & -      & -       & -        & -        \\
                           & HGAMP$_b$ vs. HALNS       & 6      & 20      & 10       & 1.59E-01 &  & 4      & 8       & 24       & 6.81E-05 \\
                           & HGAMP$_b$ vs. TBSA$_{basic}$           & 18     & 14      & 4        & 6.63E-03 &  & 22     & 2       & 12       & 4.92E-01 \\
                           & HGAMP$_b$ vs. TBSA$_{quality}$                           & 13     & 17      & 6        & 4.85E-01 &  & 11     & 4       & 21       & 7.85E-03\\
\hline
\end{tabular}
\end{center}
\end{scriptsize}
\end{table}


\begin{figure}[htbp]
\centering
\subfigure[Set $\mathbb{P}$]{
\begin{minipage}[t]{0.5\linewidth}
\centering
\includegraphics[width=3.2in]{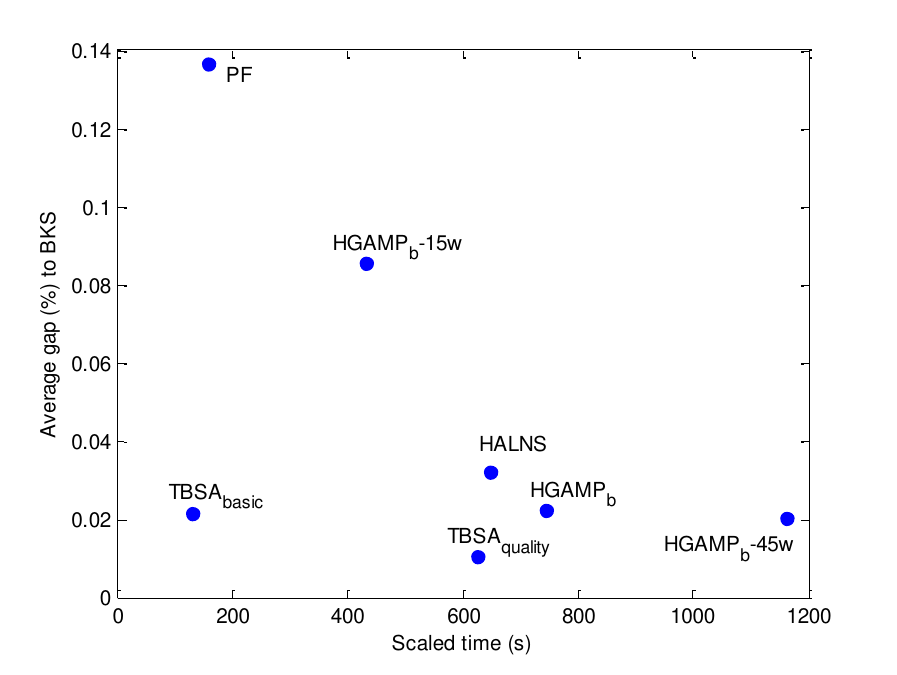}
\end{minipage}%
}%
\subfigure[Set $\mathbb{T}$]{
\begin{minipage}[t]{0.5\linewidth}
\centering
\includegraphics[width=3.2in]{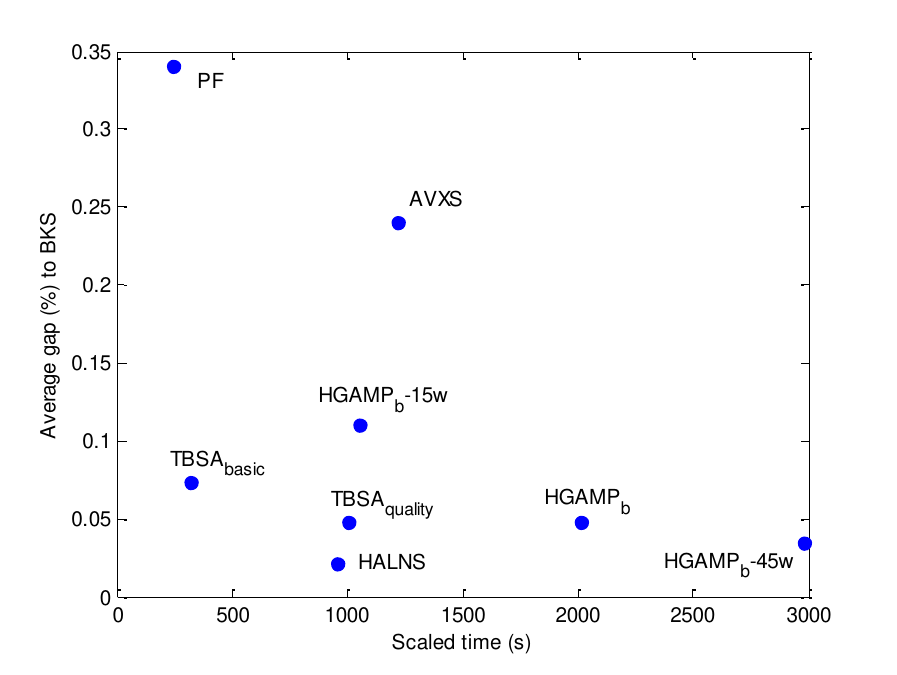}
\end{minipage}%
}%
\centering
\caption{CLRP performance comparison over the two classical sets with unknown optima.}
\label{perfor_classical}
\end{figure}


Figure \ref{perfor_classical} shows that reducing the iterations of HGAMP$_b$ by half (HGAMP$_b$-15w) results in significantly worse performance. Conversely, a slight improvement in performance can be achieved by increasing the iterations by about one-third (HGAMP$_b$-45w). Based on these findings, we have set a maximum of 300,000 iterations as the stopping condition for HGAMP. Note that HGAMP$_b$ takes relatively long to solve the set $\mathbb{T}$ because of its use of a population of over 300 individuals, which requires more time for the algorithm to converge. Fortunately, our algorithm was able to improve on three best-known results, which is noteworthy because these instances have been extensively tested by various methods and are notoriously difficult to improve on further.

\subsubsection{Performance analysis on the instance set $\mathbb{S}$}\label{new_results}

The rich and challenging instance set $\mathbb{S}$ by \citet{schneider2019large} has been used to evaluate TBSA$_{basic}$ \citep{schneider2019large}, PF \citep{arnold2021progressive} and an exact algorithm \citep{liguori2022non}. 

Table \ref{sumResults_newSet} summarizes the results and highlights the high performance of our algorithm (detailed results in Tables 3-5 of the online supplement). HGAMP$_m$ achieves a significant improvement over the BKS by obtaining 98 new upper bounds (indicated in boldface) out of the 202 instances of set $\mathbb{S}$ (49\%), whereas matching the BKS values for 26 other instances. It is worth noting that the BKS values represent the best results extracted from all existing algorithms. Compared to the highly effective heuristic TBSA$_{basic}$, HGAMP$_m$ performs much better by achieving 142 and 137 better values in terms of the best and average results, respectively. Compared to PF, HGAMP$_m$ shows a stronger dominance by obtaining 180 better results in terms of the best results. The differences between HGAMP$_m$ and its competitors are statistically significant based on the Wilcoxon signed-rank tests ($p$-values $\ll0.05$). Overall, our algorithm reaches a remarkable performance on the $\mathbb{S}$ set, and contributes to the growing body of knowledge on this challenging benchmark.

\begin{table}[h]
\caption{Summary of results between the HGAMP$_m$ and reference algorithms on set $\mathbb{S}$ of 202 instances.}\label{sumResults_newSet}
\begin{scriptsize}
\begin{center}
\begin{tabular}{p{3.4cm}p{0.9cm}p{0.9cm}p{0.9cm}p{1.5cm}p{0.01cm}p{0.9cm}p{0.9cm}p{0.9cm}p{1.5cm}}
\hline
\multicolumn{1}{c}{\multirow{2}{*}{Pair algorithms}} & \multicolumn{4}{c}{Best}               &  & \multicolumn{4}{c}{Avg.}               \\
\cline{2-5} \cline{7-10}
 \multicolumn{1}{c}{}   & \#Wins & \#Tiers & \#Losses & \textit{p}-value  &  & \#Wins & \#Tiers & \#Losses & \textit{p}-value  \\
\hline
HGAMP$_m$ vs. BKS    & \textbf{98}  & 26 & 78 & 1.73E-03 &  & -   & -  & -  & -        \\
HGAMP$_m$ vs. PF     & 180 & 6  & 16 & 4.59E-26 &  & -   & -  & -  & -        \\
HGAMP$_m$ vs. TBSA$_{basic}$   & 142 & 17 & 43 & 1.44E-11 &  & 137 & 2  & 63 & 1.01E-06 \\
HGAMP$_m$ vs. HGAMP$_b$ & 110 & 40 & 51 & 2.85E-09 &  & 127 & 11 & 64 & 3.14E-12 \\
HGAMP$_m$ vs. HGAMP$_p$ & 84  & 37 & 81 & 4.25E-01 &  & 102 & 11 & 89 & 1.26E-01\\
\hline
\end{tabular}
\end{center}
\end{scriptsize}
\end{table}


\begin{figure}[htbp]
\centering
\includegraphics[width=3.8in]{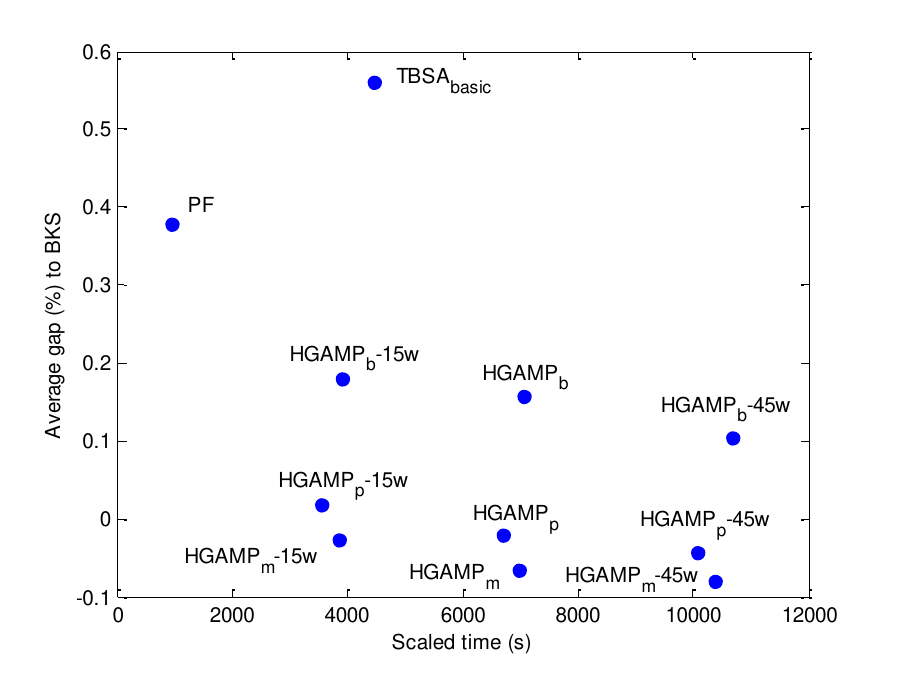}
\centering
\caption{Performance chart of the CLRP on new set $\mathbb{S}$.}
\label{perfor_new}
\end{figure}


The results presented in Figure \ref{perfor_new} demonstrate that although the scaled running time of HGAMP$_m$ is slightly longer than TBSA$_{basic}$, there is a significant performance gap between the two algorithms. A closer inspection of Tables 3-5 of the online supplement reveals that HGAMP$_m$ requires more times  to solve small instances, such as 1021.84 seconds on instance 100-5-1c, whereas TBSA$_{basic}$ can quickly find good enough results within 81.24 seconds. This difference can be attributed to the relatively large number of individuals in the population employed by HGAMP$_m$. The number of individuals in each subpopulation is approximately equal to the size of the whole population used in \cite{he2022general} and \cite{vidal2012hybrid}. As a result, HGAMP$_m$ takes more time to converge. Given a sufficient number of iterations, such as 300,000 iterations, HGAMP$_m$ performs well on both small and large instances, as shown in Tables 3-5 of the online supplement. However, TBSA$_{basic}$ needs a considerable time to solve large instances.

Table \ref{sumResults_newSet} indicates that HGAMP$_m$ significantly outperforms HGAMP$_b$ in both best and average values. Moreover, HGAMP$_m$ performs marginally better than HGAMP$_p$. These findings indicate that the depot configurations $\mathcal{D}_{pf}$ from PF are important for the performance of HGAMP$_b$. Our heuristic method, CRH, which aims to identify promising depot configurations, further enhances the performance, as evidenced by the small difference between HGAMP$_m$ and HGAMP$_p$. Figure \ref{perfor_new} also illustrates the differences between HGAMP$_m$ and the two variants. It is worth noting that HGAMP$_m$ and its two variants have essentially converged after approximately 300,000 iterations, even if HGAMP$_m$-45w achieves slightly better results on some instances. Conversely, HGAMP$_m$-15w still has the potential to significantly improve its performance.

\subsubsection{Detained analysis of the instance set $\mathbb{S}$}\label{results_analysis}

Although the $\mathbb{S}$ set has been tested by three algorithms \citep{schneider2019large,arnold2021progressive,liguori2022non}, until now, not much can be said about the quality of the solutions obtained. This section will help to fill this gap.

The 202 instances in set $\mathbb{S}$ can be categorized into four groups based on the density of customer distribution and clusters with different sizes. \citet{schneider2019large} have provided detailed descriptions of each group's characteristics. For example, instances of type 4 consist of one quarter of customers evenly distributed in the plane and the other three quarters clustered into three equally sized regions of high density. In Figure \ref{perfor_setS}(a), the five algorithms show a consistent dominance relationship across the four groups: HGAMP$_m$ $>$ HGAMP$_p$ $>$ HGAMP$_b$ $>$ PF $>$ TBSA$_{basic}$, suggesting that the algorithms' performance is not affected by variations in customer distribution and clusters with different sizes. Whereas HGAMP$_b$ yields comparable results to HGAMP$_m$ and HGAMP$_p$, it performs relatively poorly on the second and fourth groups. Conversely, PF's performance remains consistent across all four groups. However, TBSA$_{basic}$ performs worst on the first group.

\begin{figure}[htbp]
\centering
\subfigure[Different values of $t$]{
\begin{minipage}[t]{0.5\linewidth}
\centering
\includegraphics[width=3.2in]{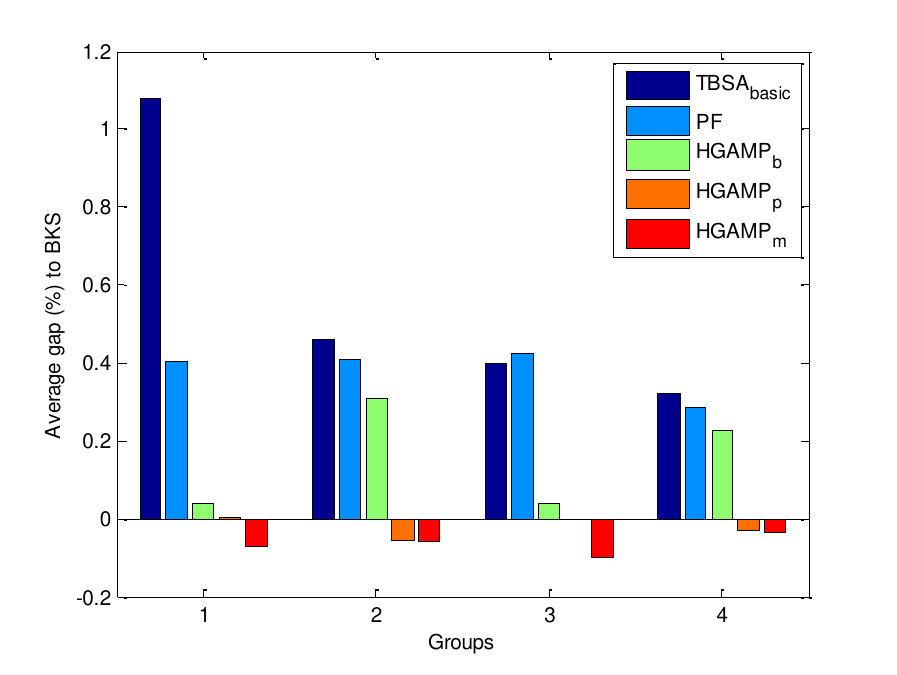}
\end{minipage}%
}%
\subfigure[Different values of $z$]{
\begin{minipage}[t]{0.5\linewidth}
\centering
\includegraphics[width=3.2in]{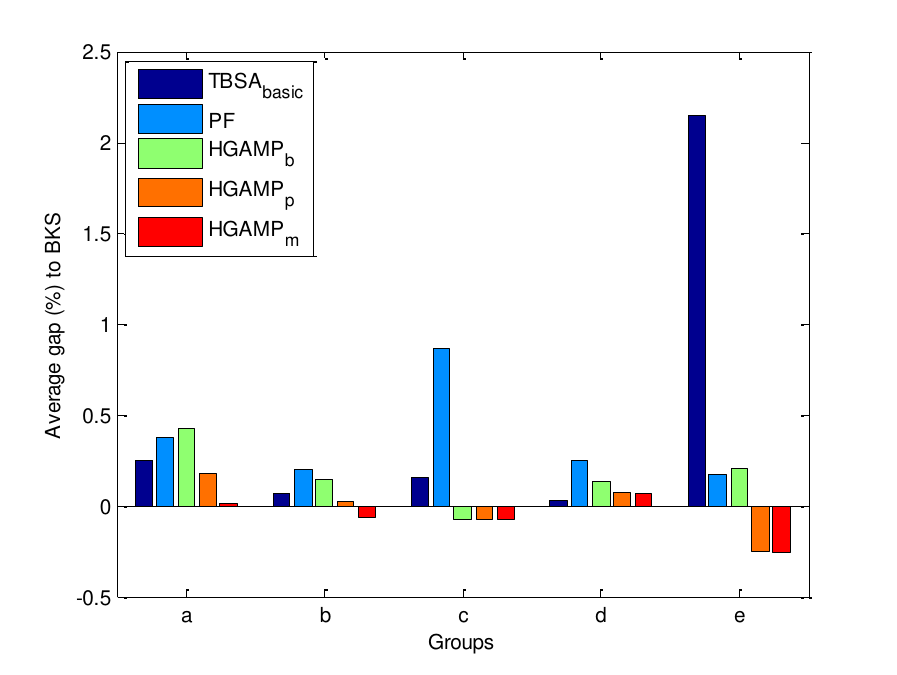}
\end{minipage}%
}%
\centering
\caption{Performance chart of the CLRP on set $\mathbb{S}$.}
\label{perfor_setS}
\end{figure}

The 202 instances in set $\mathbb{S}$ can be further classified into five groups based on vehicle capacity and depot costs. For example, type $e$ instances have larger depots with higher costs \citep{schneider2019large}. Figure \ref{perfor_setS}(b) shows that the performance of the algorithms varies significantly across the five groups, with similar performance trends for all algorithms on groups $a$, $b$, and $d$, but different trends on groups $c$ and $e$. The differences suggest that depot configurations strongly influence performance, as shown by the \textit{compact degree}, which describes the characteristics of each depot configuration. The \textit{compact degree} of instance $I$, denoted as $\omega_I = 100 \times (\sum_{j\in \mathcal{J}}d_j/{\sum_{k\in \mathcal{D}_I}w_k})$, where $\mathcal{D}_I$ is the depot configuration of the best solution of instance $I$. Then, we calculated the average \textit{compact degree} for each of the five groups: 94.70\%, 95.17\%, 27.02\%, 95.64\% and 97.62\%. Notably, groups $a$, $b$, and $d$ have similar \textit{compact degrees}, whereas groups $c$ and $e$ exhibit distinct values. For group $c$, whose \textit{compact degree} was significantly lower than the other four groups, the algorithm's performance changed greatly, and the performance of PF became extremely poor. 

Meanwhile, our three algorithms with different depot configuration construction methods showed consistent results, indicating that initial depot configurations are not so important for our HGAMP algorithm. This is because of the fact that the open cost of each depot is very small compared to the routing cost, and better solutions tend to open more depots. In other words, different depot configurations do not significantly influence the total cost. Our algorithm can construct new depot configurations from two existing depot configurations. Therefore, the initial depot configurations from the construction engine do not significantly affect the performance of HGAMP. The crossover operator in HGAMP$_m$ can effectively generate new depot configurations, ensuring that better performance can be achieved even if the initial depot configurations are not so good. Conversely, for group $e$, the use of good initial depot configurations has a decisive impact on the quality of the solution, because each depot configuration in the best solution is very compact, even 100\% in some instances. Thus, a highly effective filtering approach is helpful for the algorithm to find better results, which is also verified by Figure \ref{perfor_setS}(b). Because a highly effective filtering approach and the mdEAX crossover operator are combined in HGAMP$_m$, the algorithm can achieve highly competitive results on all five groups. In the following, we present additional experiments to investigate the contributions of the key algorithmic components to the high performance of HGAMP$_m$.

\section{Additional experiments}\label{analysis}

We now present additional experiments to study the role of the multi-population scheme and the mdEAX crossover operator. The experiments are based on the most challenging instances of the set $\mathbb{S}$.

\subsection{Rationale behind the multi-population scheme}\label{benefit_mp}

The multi-population scheme is a critical component of HGAMP as it enables the management of population  solutions, the maintenance of population diversity, and the updating of promising depot configurations. To evaluate the effectiveness of this scheme, we compared three variants of HGAMP$_m$ with different values of $\gamma$ (indicating $\gamma$+1 subpopulations): HGAMP$_0$ ($\gamma = 0$, corresponding to a single population), HGAMP$_5$ ($\gamma = 5$), and HGAMP$_{15}$ ($\gamma = 15$), along with HGAMP$_m$ where $\gamma = 10$. To ensure a fair comparison, we ran all compared variants on the same machine with the same parameters as HGAMP$_m$. The results of the comparison are summarized in Table \ref{compar_mp}.

\begin{table}[h]
\caption{Summary of comparative results of HGAMP$_m$ with three variants with the different number of subpopulations.}\label{compar_mp}
\begin{scriptsize}

\begin{center}
\begin{tabular}{p{4.0cm}p{0.8cm}p{0.9cm}p{0.9cm}p{1.5cm}p{0.01cm}p{0.8cm}p{0.9cm}p{0.9cm}p{1.5cm}}
\hline
\multicolumn{1}{c}{\multirow{2}{*}{Pair algorithms}} & \multicolumn{4}{c}{Best}                       &  & \multicolumn{4}{c}{Avg.}                       \\
\cline{2-5}\cline{7-10}
\multicolumn{1}{c}{}     & \#Wins & \#Tiers & \#Losses & \textit{p}-value &  & \#Wins & \#Tiers & \#Losses & \textit{p}-value \\
\hline
HGAMP$_0$ vs HGAMP$_m$        & 19     & 28      & 155       & 2.65E-24         &  & 8     & 4      & 190       & 2.90E-33         \\
HGAMP$_5$ vs HGAMP$_m$        & 59     & 31      & 112       & 6.20E-04         &  & 32     & 8      & 162       & 6.31E-23         \\
HGAMP$_{15}$ vs HGAMP$_m$        & 92     & 34      & 76       & 1.03E-01         &  & 135     & 9      & 58       & 2.98E-09         \\
\hline
\end{tabular}
\end{center}
\end{scriptsize}
\end{table}

Table \ref{compar_mp} shows that HGAMP$_m$ significantly outperforms HGAMP$_0$ and HGAMP$_5$ in terms of both the best and average values, whereas HGAMP$_{15}$ is slightly better than HGAMP$_m$ for best values and significantly outperforms HGAMP$_m$ for average values. these results indicate that a larger number of subpopulations leads to performance improvement. However, as the number of subpopulations increases further, the performance improvement gradually becomes marginal. In conclusion, the multi-population scheme is a critical component of HGAMP that enhances its performance. 


\subsection{Rationale behind the crossover}\label{benefit_mdEAX}

To evaluate the effectiveness of the mdEAX crossover, we created two HGAMP$_m$ variants. The first variant, HGAMP$_d$, replaces mdEAX with the \textit{ChangeDepot} operator. This operator removes a depot from the current solution and then greedily reinserts the unvisited customers back into the solution. Specifically, we select a solution $\varphi$ from $\mathcal{P}$, and then choose a depot $i\in\mathcal{D}_{\varphi}$ based on the minimum capacity utility $T_i/{w_i}$, where $T_i = \sum_{j\in \mathcal{S}_i}{d_j}$ and $\mathcal{S}_i$ is the set of customers covered by depot $i$ in solution $\varphi$. Next, all customers $j\in \mathcal{S}_i$ are removed from the current solution and depot $i$ is removed from $\mathcal{D}_{\varphi}$. Then, the customers from set $\mathcal{S}_i$ are greedily reinserted back into the partial solution $\varphi$ with respect to the objective. The second variant, HGAMP$_c$, replaces the mdEAX crossover with two neighborhood operators: \textit{ChangeDepot} and the mutation operator. In HGAMP$_c$, we randomly use one of the two operators to generate new solutions in each generation. To ensure a fair comparison, we ran these variants on the same machine with the same parameters as HGAMP$_m$. We summarize the results of this comparison in Table \ref{compar_cross}.

\begin{table}[h]
\caption{Summary of comparative results of HGAMP$_m$ with two variants.}\label{compar_cross}
\begin{scriptsize}
\begin{center}
\begin{tabular}{p{3.9cm}p{1.1cm}p{1.1cm}p{1.1cm}p{1.5cm}p{0.01cm}p{1.1cm}p{1.1cm}p{1.1cm}p{1.5cm}}
\hline
\multicolumn{1}{c}{\multirow{2}{*}{Pair algorithms}} & \multicolumn{4}{c}{Best}                       &  & \multicolumn{4}{c}{Avg.}                       \\
\cline{2-5}\cline{7-10}
\multicolumn{1}{c}{}     & \#Wins & \#Tiers & \#Losses & \textit{p}-value &  & \#Wins & \#Tiers & \#Losses & \textit{p}-value \\
\hline
HGAMP$_d$ vs HGAMP$_m$        & 0     & 10      & 192       & 2.94E-33         &  & 1     & 1      & 200       & 1.54E-34         \\
HGAMP$_c$ vs HGAMP$_m$        & 1     & 15      & 186       & 2.11E-32         &  & 2     & 2      & 198       & 4.70E-34         \\
\hline
\end{tabular}
\end{center}
\end{scriptsize}
\end{table}

Table \ref{compar_cross} clearly shows that HGAMP$_m$ outperforms HGAMP$_d$ in terms of both the best and average values, with HGAMP$_m$ achieving better results in terms of the average values in all but one case. Moreover, when the mutation operator is introduced in HGAMP$_c$, the results show a slight improvement, with HGAMP$_c$ only achieving 1 and 2 better solutions in terms of the best and average results. 


\section{Conclusions}\label{conclusion}

In this paper, we propose a hybrid genetic algorithm with multi-population (HGAMP) for the capacitated location-routing problem. This approach maintains multiple subpopulations where each subpopulation consists of a set of high-quality solutions that share the same depot configuration. The algorithm uses a multi-depot edge assembly crossover (mdEAX) to explore promising depot configurations and an effective neighborhood-based local optimization to perform route optimization. The algorithm additionally applies a coverage ratio heuristic for initial depot configuration generation and a special  mutation to enhance diversity.

Computational experiments on four sets of 281 commonly used benchmark instances show that the proposed algorithm performs remarkably well, finding 101 new best-known results (improved upper bounds) and matching the best-known results for 84 other instances. These new results can be valuable for future research on the problem, especially in assessing the performance of new algorithms on instances with different characteristics, such as customer distribution density and depot costs. In addition, we investigate the role and rationale of the multi-population scheme as well as the mdEAX operator for the CLRP. Because the CLRP is a relevant model for a number of real-world problems, our algorithm, whose code will be publicly available, can be used to better solve some of these practical applications.

Beyond the CLRP studied in this paper, the proposed framework using multiple populations and multi-depot edge assembly crossover could be helpful for better solving related location routing problems where both location and routing decisions need to be made, such as periodic location routing, two-echelon vehicle routing and location arc routing. In fact, it is natural to adapt the main idea of our approach, i.e., simultaneous exploration of unknown depot configurations and new route solutions with edge assembly, to bridge the gaps encountered in the context of these location routing problems. 


\section*{Acknowledgments}
The authors would like to thank Dr. F. Arnold and Prof. K. S{\"o}rensen for kindly sharing their source code of the PF algorithm \citep{arnold2021progressive}. This work was partially supported by the National Natural Science Foundation of China under Grant No. 72122006.

\bibliographystyle{informs2014}
\bibliography{ma_clrp} 

\begin{APPENDIX}{Online Supplement}

%
%
%

\section{Coverage ratio heuristic}\label{cover_heuristic}
CRH is composed of two successive filters. During the preliminary filter, we rapidly construct many depot configurations, taking into account the best balance between roughly estimated costs and geographic dispersion. In the secondary filter, we narrow down the options and identify the most promising depot configurations by examining a small number of complete solutions.

\subsection{Preliminary filter}\label{pre_fliter}
\begin{algorithm}[H]\label{algo:first}
\renewcommand{\baselinestretch}{0.6}\huge\normalsize
\begin{scriptsize}
\caption{The preliminary filter heuristic for depot configurations}
\DontPrintSemicolon 
\KwIn{Instance $I$, parameters $r_{min}$, $r_{max}$, $H_{max}$ and $I_{max}$;}
\KwOut{A set of depot configurations $\mathcal{D}$;}
\Begin{
\For{$i=1\ to\ |\mathcal{I}|$}
{
	$\{\mathcal{S}_i,u_i\} \gets MST(I,i)$;\tcc*[l]{Use the MST to evaluate the cost of each depot}
}
$\mathcal{D}\gets \varnothing$;\\
$m \gets 0$;\\
\While{$|\mathcal{D}| <$ $H_{max}$ and $m < I_{max}$}{
	 A random depot $s$ is targeted and $\mathcal{H}_c \gets \{s\}$;\\
	 $\mathcal{S}_{\mathcal{H}}\gets \mathcal{S}_s$;\\
	 $T_c \gets w_s$;\\
	\While{$T_c < T$}{

		$\mathcal{L} \gets \varnothing$;\tcc*[l]{A list to save all candidate depots}	
		\For{$i =1\ to\  |\mathcal{I} \backslash \mathcal{H}|$ }{
		$r_i \gets |\mathcal{S}_{\mathcal{H}} \cap \mathcal{S}_i| / |\mathcal{S}_{\mathcal{H}} \cup \mathcal{S}_i|$;\\
		$r_{b} \gets (r_{max}-r_{min}) \times ((w_i+T_c)/T) + r_{min}$;\\
	 	\If{$r_i< r_{b}$}{
	 		$\mathcal{L} \gets \mathcal{L}\cup \{u_i, i\}$;\tcc*[l]{Sort each depot with respect to $u_i$}
	 	}
	}
		\If{$\mathcal{L} = \varnothing$}{
		$\mathcal{H} \gets \varnothing$;\\
		\textit{Break};\tcc*[l]{Cannot find suitable depots}
	}
	
	$i \gets P(\mathcal{L})$;\tcc*[l]{Find a depot from list $\mathcal{L}$ with probability}
	$\mathcal{S}_{\mathcal{H}}\gets \mathcal{S}_{\mathcal{H}} \cup \mathcal{S}_i$;\\
	$\mathcal{H} \gets \mathcal{H} \cup \{i\}$;\\
	$T_c \gets T_c + w_i$;\\
	}
	\eIf{$\mathcal{H} = \varnothing$}{
		$m \gets m+1$;\\
	}
	{
		$\mathcal{D} \gets \mathcal{D} \cup \{\mathcal{H}\}$;\\
		$m \gets 0$;\\
	}
}
\Return{$\mathcal{D}$};\\
}
\end{scriptsize}
\end{algorithm}
\renewcommand{\baselinestretch}{0.6}\huge\normalsize
Before constructing depot configurations, it is necessary to estimate the approximate costs associated with each depot configuration that will be involved in the complete solutions. To achieve this, one commonly employed method is based on the minimum spanning tree (MST) determined by Prim's algorithm.  Prim's algorithm starts with a root node ($i \in \mathcal{I}$), as shown in Algorithm \ref{algo:first}, lines 2-4, and constructs a MST based on its capacity $w_i$. Specifically, edges emanating from the root node are extended, and the corresponding customers are saved in set $\mathcal{S}_i$. Unvisited customers are continually evaluated, and they are added into $\mathcal{S}_i$ if the total demand of the visited customers is less than $w_i$. The rough cost ($u_i$) of depot $i$ is determined as the sum of the open cost $o_i$, the fixed costs of the vehicles required to meet the accumulated demand of the customers in $\mathcal{S}_i$, and the total travel cost of the MST. The set $\mathcal{S}_i$ can be thought of as the coverage area for depot $i$.

\begin{figure}[htbp]
\centering
\includegraphics[width=3.5in]{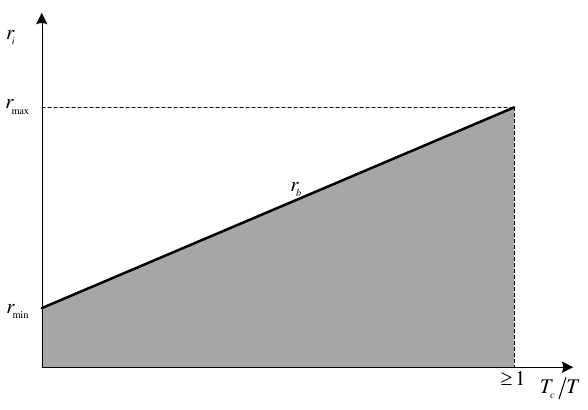}
\centering
\caption{Illustration of the overlap coverage ratio}
\label{fig_crh}
\end{figure}


Intuitively, balancing approximate costs and geographic dispersion is essential for identifying promising depot configurations that minimize depot costs and avoid repeatedly covering the same customers. In Algorithm \ref{algo:first},  $T = \sum_{j\in \mathcal{J}}d_j$ is the total demand of all customers, $r_{min}$ and $r_{max}$ are the minimum and maximum overlap coverage ratio, respectively. The heuristic also imposes a maximum limit of depot configurations ($H_{max}$) and iterations ($I_{max}$) to terminate the search.

In Algorithm \ref{algo:first}, a random depot is used to initialize set $\mathcal{H}$ (line 8). The overlap coverage ratio $r_i$ of depot $i$ is then calculated as the proportion of customers that overlap between $\mathcal{S}_{\mathcal{H}}$ and $\mathcal{S}_i$ (line 14). Ideally, the overlap ratio should be as small as possible to avoid that the selected depots cover many identical customers. Figure \ref{fig_crh} shows that as the total capacity ($T_c$) of the selected depots increases, the requirement for comparative overlap ratios is gradually relaxed as more and more of the same customers are covered by $\mathcal{S}_{\mathcal{H}}$ and $\mathcal{S}_i$. Each candidate depot ($r_i < r_b$) is added to a list $\mathcal{L}$, sorted based on approximate costs. Subsequently, a depot is selected from $\mathcal{L}$ with a probability (line 24) and is added to the current depot configuration $\mathcal{H}$. We use a parameter $p_d$ $(p_d>1)$ to introduce some randomness in the selection process, giving higher probability to the first depot. Specifically, let $r = y^{p_d}*|\mathcal{L}|$, where $y$ is a random real number in $(0,1)$. Then, the $r$th element in the list $\mathcal{L}$ is selected as the candidate depot and added to $\mathcal{H}$. The process is repeated until a feasible depot configuration ($T_c \geq T$) is constructed and then we move on to create the next depot configuration. Without loss of generality, all parameters in Algorithm \ref{algo:first} are independent of the instance size. We set $r_{min} = 0.1$, $r_{max} = 0.6$, $H_{max} = 1000$, $I_{max} = 1000$, and $p_d = 6$. Based on our preliminary tests, the running time of Algorithm \ref{algo:first} is negligible. 

Of particular interest is the gradual increase in the total capacity of the currently selected depots and the relaxed constraint on the overlap coverage ratio, which provides more flexibility in identifying promising depot configurations. However, focusing too much on geographic dispersion can have an irreparable impact on the fixed costs of depots, while ignoring routing costs is also problematic. Therefore, the preliminary filter aims to balance these two aspects to construct as many depot configurations as possible. Since the size of set $\mathcal{D}$ remains large, the secondary filter in the next section is necessary to identify a reduced set of final depot configurations.

\subsection{Secondary filter}{

After applying the preliminary filter, there remain many possible depot configurations. However, some of these configurations may not be promising because the MST can only provide approximate costs per depot involved in complete solutions. Therefore, a more fine-grained filtering process is necessary to identify the most promising depot configurations. 

\renewcommand{\baselinestretch}{0.8}\huge\normalsize
\begin{algorithm}[H]\label{algo:second}
\begin{scriptsize}
\caption{The secondary filter heuristic for depot configurations}
\DontPrintSemicolon 
\KwIn{Instance $I$, set $\mathcal{D}$, parameters $N_t$, $\gamma$;}
\KwOut{A set of promising depot configurations $\mathcal{D}$;}
\Begin{
$\mathcal{L} \gets \varnothing$;\\
\For{$i=1\ to\ |\mathcal{D}|$}
{
	$\mathcal{B}_i \gets \varnothing$;\\
	\For{$j=1\ to\ N_t$}{
		$\varphi_j \gets $\textit{RGH}($\mathcal{D}_i, I$);\tcc*[l]{Using random greedy heuristic to construct each solution, Section 3.2.1}
		$\varphi_j \gets $\textit{LocalSearch}($\varphi_j$);\tcc*[l]{Improve each solution, Section 3.2.5}
		$\mathcal{B}_i \gets \mathcal{B}_i \cup \{\varphi_j\}$;\\
	}
	$a_i \gets \frac{\sum_{i=1}^{|\mathcal{B}_i|}{f(\varphi_i)}}{|\mathcal{B}_i|}$ ;\tcc*[l]{Obtaining average costs}
	$\mathcal{L} \gets \mathcal{L}\cup \{a_i,\mathcal{D}_i\}$;\tcc*[l]{Sort each depot configuration with respect to $a_i$}
}
First $\gamma$ depot configurations from $\mathcal{L}$ are kept in set $\mathcal{D}$;\\
\Return{$\mathcal{D}$};\\
}
\end{scriptsize}
\end{algorithm}
\renewcommand{\baselinestretch}{1.0}\huge\normalsize
}

As shown in Algorithm \ref{algo:second}, for each depot configuration $\mathcal{D}_i$, a random greedy heuristic algorithm (RGH) constructs a number of initial solutions, which are then improved by the local search procedure of Section 3.2.5. The improved solutions are inserted into a corresponding pool $\mathcal{B}_i$, which is associated with the corresponding depot configuration $\mathcal{D}_i$. However, due to the large number of candidate configurations and the non-negligible running time of each local search, the number of solutions in the pool of depot configurations is limited to $N_t = 10$. The average cost of each pool is calculated to determine whether the corresponding depot configuration is promising, and then all depot configurations are sorted based on their average costs. Finally, the top $\gamma$ depot configurations, where $\gamma = 10$, are selected as the initial depot configurations for the proposed HGAMP algorithm.

\section{Computational results}\label{comput_results}

This section presents the detailed computational results of the proposed HGAMP algorithm together with the results of the reference algorithms PF \citep{arnold2021progressive}, HALNS \citep{voigt2022hybrid}, TBSA\_{basic} \citep{schneider2019large} and TBSA\_{quality} \citep{schneider2019large}. In Tables \ref{tableP}-\ref{tableS3},  column Instance indicates the name of each instance; column BKS shows the best-known results (best upper bounds) summarized from the literature, including both exact algorithms \citep{liguori2022non,contardo2014exact} and representative heuristics \citep{arnold2021progressive,voigt2022hybrid,schneider2019large}; Best and Avg. are the best and average results over 20 independent runs obtained by the corresponding algorithm in the column header, respectively; Time in each column is the average running time in seconds of the corresponding algorithm; for HGAMP, column TMB is the average time in seconds needed by HGAMP to attain its best results (thus TMB $\leq$ Time always holds); $\delta$(\%) in the last column is calculated as $\delta=100 \times (f_{best}-BKS)/BKS$, where $f_{best}$ is the best objective value of HGAMP. The \textit{Average} row is the average value of a performance indicator over the instances of a benchmark set. Improved best results (new upper bounds) are indicated by negative $\delta$(\%) values highlighted in boldface. In all tables, except Set $\mathbb{B}$ in Table \ref{tableP} since most algorithms obtain the same value for each instance, the dark gray color indicates that the corresponding algorithm obtains the best result among the compared algorithms on the corresponding instance; the medium gray color displays the second best results, and so on.

       \begin{rotate}
      \begin{table}[!h]
      \renewcommand{\arraystretch}{1.5}
      \centering
      \TABLE
      {Results for the CLRP on the instances of set $\mathbb{P}$. \label{tableP}}
      {\begin{tiny}
\begin{tabular}{p{0.3cm}p{2.0cm}p{0.9cm}p{0.9cm}p{0.4cm}p{0.01cm}p{0.9cm}p{1.0cm}p{0.4cm}p{0.01cm}p{0.9cm}p{1.0cm}p{0.4cm}p{0.01cm}p{0.9cm}p{1.0cm}p{0.4cm}p{0.01cm}p{0.9cm}p{1.0cm}p{0.7cm}p{0.7cm}p{0.7cm}}

\hline
& &  & \multicolumn{2}{c}{PF}                  &  & \multicolumn{3}{c}{HALNS}                      &  & \multicolumn{3}{c}{TBSA$_{basic}$}  &  & \multicolumn{3}{c}{TBSA$_{quality}$}                    &  & \multicolumn{5}{c}{HGAMP$_b$}                                              \\
\cline{4-5} \cline{7-9} \cline{11-13} \cline{15-17} \cline{19-23}
\multirow{-2}{*}{Set}&\multirow{-2}{*}{Instances} & \multirow{-2}{*}{BKS} & Best                           & Time   &  & Best                           & Avg.      & Time    &  & Best                           & Avg.      & Time   &  & Best                           & Avg.      & Time    &  & Best                           & Avg.      & Time    & TMB     & $\delta$(\%)   \\
\hline
                         & 20-5-1a                                         & 54793                                     & \cellcolor[HTML]{A6A6A6}54793  & 20.00  &  & \cellcolor[HTML]{A6A6A6}54793  & 54793.00  & 25.61   &  & \cellcolor[HTML]{A6A6A6}54793  & 54793.00  & 2.97   &  & \cellcolor[HTML]{A6A6A6}54793  & 54793.00  & 15.84   &  & \cellcolor[HTML]{A6A6A6}54793  & 54793.00  & 3.54    & 0.39    & 0.00           \\
                         & 20-5-1b                                         & 39104                                     & \cellcolor[HTML]{A6A6A6}39104  & 26.00  &  & \cellcolor[HTML]{A6A6A6}39104  & 39104.00  & 21.30   &  & \cellcolor[HTML]{A6A6A6}39104  & 39104.00  & 2.04   &  & \cellcolor[HTML]{A6A6A6}39104  & 39104.00  & 11.46   &  & \cellcolor[HTML]{A6A6A6}39104  & 39104.00  & 2.56    & 0.90    & 0.00           \\
\hline    
\end{tabular}
\end{tiny}
}

      \end{table}%
      \end{rotate}

      \begin{rotate}
      \begin{table}[!h]
            \renewcommand{\arraystretch}{1.1}
      \centering
      \TABLE
      {Results for the CLRP on the instances of set $\mathbb{T}$. \label{tableT}}
      {\begin{tiny}
     \begin{tabular}{p{0.2cm}p{0.8cm}p{0.7cm}p{0.7cm}p{0.4cm}p{0.01cm}p{0.7cm}p{0.4cm}p{0.01cm}p{0.7cm}p{0.7cm}p{0.4cm}p{0.01cm}p{0.7cm}p{0.7cm}p{0.4cm}p{0.01cm}p{0.7cm}p{0.7cm}p{0.4cm}p{0.01cm}p{0.7cm}p{0.7cm}p{0.7cm}p{0.7cm}p{0.7cm}}
\hline
\multicolumn{1}{c}{}                            & \multicolumn{1}{c}{}                      & \multicolumn{3}{c}{AVXS}                            &  & \multicolumn{2}{c}{PF}                   &  & \multicolumn{3}{c}{HALNS}                           &  & \multicolumn{3}{c}{TBSA$_{basic}$}                    &  & \multicolumn{3}{c}{TBSA$_{quality}$}          &  & \multicolumn{4}{c}{HGAMP$_b$}                                     &                \\
\cline{3-5} \cline{7-8} \cline{10-12} \cline{14-16} \cline{18-20} \cline{22-26}
\multicolumn{1}{c}{\multirow{-2}{*}{Instances}} & \multicolumn{1}{c}{\multirow{-2}{*}{BKS}} & Best                            & Avg.    & Time    &  & Best                            & Time   &  & Best                            & Avg.    & Time    &  & Best                            & Avg.    & Time   &  & Best                            & Avg.    & Time    &  & Best                            & Avg.    & Time    & TMB     & $\delta$(\%)            \\
\hline
P111112                                         & 1467.68                                   & \cellcolor[HTML]{808080}1467.68 & 1467.68 & 110.00  &  & 1468.29                         & 92.00  &  & \cellcolor[HTML]{808080}1467.68 & 1467.68 & 163.22  &  & \cellcolor[HTML]{808080}1467.68 & 1467.80 & 101.07 &  & \cellcolor[HTML]{808080}1467.68 & 1467.68 & 345.07  &  & \cellcolor[HTML]{808080}1467.68 & 1467.68 & 1268.00 & 1193.14 & 0.00           \\
\hline               
\end{tabular}
\end{tiny}
}

      \end{table}%
      \end{rotate}%

      \begin{rotate}
      \begin{table}[!h]
      \renewcommand{\arraystretch}{1.2}
      \centering
      \TABLE
      {Results for the CLRP on the instances of set $\mathbb{S}$. \label{tableS1}}
      {\begin{tiny}
    \begin{tabular}{p{1.2cm}p{0.8cm}p{0.8cm}p{1.0cm}p{0.6cm}p{0.01cm}p{0.8cm}p{0.4cm}p{0.01cm}p{0.8cm}p{1.0cm}p{0.6cm}p{0.01cm}p{0.8cm}p{1.0cm}p{0.6cm}p{0.01cm}p{0.8cm}p{1.0cm}p{0.8cm}p{0.8cm}p{0.7cm}}
\hline
\multicolumn{1}{c}{}  & \multicolumn{1}{c}{} & \multicolumn{3}{c}{TBSA$_{basic}$}     &  & \multicolumn{2}{c}{PF}  &  & \multicolumn{3}{c}{HGAMP$_b$} &  & \multicolumn{3}{c}{HGAMP$_p$}     &  & \multicolumn{5}{c}{HGAMP$_m$}    \\
\cline{3-5} \cline{7-8} \cline{10-12} \cline{14-16} \cline{18-22}
\multicolumn{1}{c}{\multirow{-2}{*}{Instances}} & \multicolumn{1}{c}{\multirow{-2}{*}{BKS}} & Best     & Avg.      & Time   &  & Best                          & Time &  & Best    & Avg.      & Time    &  & Best    & Avg.      & Time    &  & Best  & Avg.      & Time    & TMB     &  $\delta$(\%)               \\
\hline
100-5-1c                                        & 134516                                    & \cellcolor[HTML]{A6A6A6}134516 & 134603.80 & 81.24  &  & 134687                         & 112  &  & \cellcolor[HTML]{A6A6A6}134516 & 134516.00 & 1024.66 &  & \cellcolor[HTML]{A6A6A6}134516 & 134516.00 & 1058.84 &  & \cellcolor[HTML]{A6A6A6}134516 & 134516.00 & 1021.84 & 946.35  & 0.00           \\
100-5-1d                                        & 275749                                    & \cellcolor[HTML]{A6A6A6}275749 & 275793.40 & 52.71  &  & 276154                         & 122  &  & \cellcolor[HTML]{A6A6A6}275749 & 275751.85 & 982.13  &  & \cellcolor[HTML]{A6A6A6}275749 & 275757.45 & 995.65  &  & \cellcolor[HTML]{A6A6A6}275749 & 275751.85 & 976.36  & 885.19  & 0.00           \\
100-5-1e                                        & 292301                                    & \cellcolor[HTML]{A6A6A6}292311 & 292400.40 & 68.68  &  & 292565                         & 159  &  & \cellcolor[HTML]{A6A6A6}292311 & 293097.10 & 1151.87 &  & \cellcolor[HTML]{A6A6A6}292311 & 293163.50 & 1083.20 &  & \cellcolor[HTML]{A6A6A6}292311 & 292941.15 & 1068.39 & 972.10  & 0.00           \\
\hline               
\end{tabular}
\end{tiny}
}

      \end{table}%
      \end{rotate}%

      \begin{rotate}
      \begin{table}[!h]
      \renewcommand{\arraystretch}{1.2}
      \centering
      \TABLE
      {Results for the CLRP on the instances of set $\mathbb{S}$. \label{tableS2}}
      {\begin{tiny}
  \begin{tabular}{p{1.2cm}p{0.8cm}p{0.8cm}p{1.1cm}p{0.6cm}p{0.01cm}p{0.8cm}p{0.4cm}p{0.01cm}p{0.8cm}p{1.1cm}p{0.6cm}p{0.01cm}p{0.8cm}p{1.1cm}p{0.6cm}p{0.01cm}p{0.8cm}p{1.1cm}p{0.8cm}p{0.8cm}p{0.7cm}}
\hline
\multicolumn{1}{c}{}  & \multicolumn{1}{c}{} & \multicolumn{3}{c}{TBSA$_{basic}$}     &  & \multicolumn{2}{c}{PF}  &  & \multicolumn{3}{c}{HGAMP$_b$} &  & \multicolumn{3}{c}{HGAMP$_p$}     &  & \multicolumn{5}{c}{HGAMP$_m$}    \\
\cline{3-5} \cline{7-8} \cline{10-12} \cline{14-16} \cline{18-22}
\multicolumn{1}{c}{\multirow{-2}{*}{Instances}} & \multicolumn{1}{c}{\multirow{-2}{*}{BKS}} & Best     & Avg.      & Time   &  & Best                          & Time &  & Best    & Avg.      & Time    &  & Best    & Avg.      & Time    &  & Best  & Avg.      & Time    & TMB     &  $\delta$(\%)               \\
\hline
300-15-2a                                       & 757931                                    & \cellcolor[HTML]{BFBFBF}759999  & 760278.40  & 2135.43 &  & 762089                          & 367  &  & \cellcolor[HTML]{D9D9D9}760052  & 760642.50  & 4800.43  &  & \cellcolor[HTML]{A6A6A6}758289  & 760344.15  & 5100.32  &  & \cellcolor[HTML]{F2F2F2}760167  & 761096.15  & 5114.15  & 4849.90  & 0.30           \\
300-15-2b                                       & 557525                                    & \cellcolor[HTML]{F2F2F2}557912  & 559174.80  & 1250.03 &  & \cellcolor[HTML]{A6A6A6}557525  & 366  &  & \cellcolor[HTML]{BFBFBF}556945  & 557703.05  & 4728.11  &  & \cellcolor[HTML]{D9D9D9}557075  & 557676.80  & 4725.22  &  & \cellcolor[HTML]{A6A6A6}556938  & 557464.40  & 4718.80  & 4385.91  & \textbf{-0.11} \\
\hline               
\end{tabular}
\end{tiny}
}

      \end{table}%
      \end{rotate}%

     \begin{rotate}
      \begin{table}[!h]
      \renewcommand{\arraystretch}{1.2}
      \centering
      \TABLE
      {Results for the CLRP on the instances of set $\mathbb{S}$. \label{tableS3}}
      {\begin{tiny}
  \begin{tabular}{p{1.1cm}p{1.1cm}p{1.1cm}p{1.1cm}p{0.7cm}p{0.01cm}p{1.1cm}p{0.4cm}p{0.01cm}p{1.1cm}p{1.1cm}p{0.7cm}p{0.01cm}p{1.1cm}p{1.1cm}p{0.7cm}p{0.01cm}p{1.1cm}p{1.1cm}p{0.8cm}p{0.8cm}p{0.7cm}}
\hline
\multicolumn{1}{c}{}  & \multicolumn{1}{c}{} & \multicolumn{3}{c}{TBSA$_{basic}$}     &  & \multicolumn{2}{c}{PF}  &  & \multicolumn{3}{c}{HGAMP$_b$} &  & \multicolumn{3}{c}{HGAMP$_p$}     &  & \multicolumn{5}{c}{HGAMP$_m$}    \\
\cline{3-5} \cline{7-8} \cline{10-12} \cline{14-16} \cline{18-22}
\multicolumn{1}{c}{\multirow{-2}{*}{Instances}} & \multicolumn{1}{c}{\multirow{-2}{*}{BKS}} & Best     & Avg.      & Time   &  & Best                          & Time &  & Best    & Avg.      & Time    &  & Best    & Avg.      & Time    &  & Best  & Avg.      & Time    & TMB     & $\delta$(\%)            \\
\hline
400-25-4a                                       & 1015654                                   & \cellcolor[HTML]{F2F2F2}1015654 & 1016764.20 & 5539.99 &  & 1016670                         & 546  &  & \cellcolor[HTML]{D9D9D9}1015020 & 1017262.00 & 8342.61  &  & \cellcolor[HTML]{BFBFBF}1014820 & 1016848.00 & 7951.03  &  & \cellcolor[HTML]{A6A6A6}1014790 & 1016562.00 & 8112.45  & 7562.81  & \textbf{-0.09} \\
400-25-4b                                       & 801722                                    & \cellcolor[HTML]{F2F2F2}801722  & 803138.40  & 3117.51 &  & 803237                          & 532  &  & \cellcolor[HTML]{D9D9D9}801700  & 802640.85  & 8977.46  &  & \cellcolor[HTML]{A6A6A6}800709  & 802041.90  & 8285.89  &  & \cellcolor[HTML]{BFBFBF}801304  & 801993.90  & 8456.36  & 7876.49  & \textbf{-0.05} \\
\hline               
\end{tabular}
\end{tiny}
}

      \end{table}%
      \end{rotate}%

\end{APPENDIX}

\end{document}